\documentclass[sn-chicago,review]{sn-jnl}% Chicago-based Humanities Reference Style

\usepackage{enumerate}
\usepackage{graphicx}%
\usepackage{multirow}%
\usepackage{amsmath,amssymb,amsfonts}%
\usepackage{amsthm}%
\usepackage{mathrsfs}%
\usepackage[title]{appendix}%
\usepackage{xcolor}%
\usepackage{textcomp}%
\usepackage{manyfoot}%
\usepackage{booktabs}%
\usepackage{algorithm}%
\usepackage{algorithmicx}%
\usepackage{algpseudocode}%
\usepackage{listings}%
\usepackage{listing}
\usepackage{cleveref}%
\usepackage{mathpartir}

\theoremstyle{thmstyleone}%
\theoremstyle{thmstyletwo}%

\theoremstyle{thmstylethree}%
\newtheorem{definition}{Definition}%

\begin{document}
\renewcommand{\orcidlogo}{%
  \includegraphics[width=10pt]{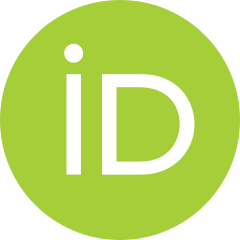}%
}
\renewcommand{\orcid}[1]{\href{https://orcid.org/#1}{\orcidlogo}}

\title[GDPR-based Access Control in Distributed Systems]{Lawful and Accountable Personal Data Processing with GDPR-based Access and Usage Control in Distributed Systems (Under Submission)}


\author*[1]{\fnm{L. Thomas} \sur{van Binsbergen}\orcid{0000-0001-8113-2221}}\email{ltvanbinsbergen@acm.org}

\author[1]{\fnm{Marten C.} \sur{Steketee}\orcid{0009-0007-8297-1909}}\email{m.c.steketee@uva.nl}

\author[1]{\fnm{Milen G.} \sur{ Kebede}\orcid{0000-0003-4790-7024}}\email{m.g.kebede@uva.nl}

\author[2]{\fnm{Heleen L.} \sur{ Janssen}\orcid{0000-0002-2785-5741}}\email{h.l.janssen@uva.nl}

\author[1,2,3]{\fnm{Tom M.} \sur{van Engers}\orcid{0000-0003-3699-8303}}\email{vanengers@uva.nl}

\affil[1]{\orgdiv{Informatics Institute}, \orgname{University of Amsterdam}, %\orgaddress{\street{Street}, \city{City}, \postcode{100190}, \state{State}, \country{Country}}}
\orgaddress{\city{Amsterdam}, \country{The Netherlands}}}

\affil[2]{\orgdiv{Institute for Information Law}, \orgname{University of Amsterdam}, %\orgaddress{\street{Street}, \city{City}, \postcode{100190}, \state{State}, \country{Country}}}
\orgaddress{\city{Amsterdam}, \country{The Netherlands}}}

\affil[3]{\orgdiv{Leibniz Institute}, \orgname{TNO}, %\orgaddress{\street{Street}, \city{City}, \postcode{100190}, \state{State}, \country{Country}}}
\orgaddress{\city{Amsterdam/The Hague}, \country{The Netherlands}}}


\abstract{
Compliance with the GDPR privacy regulation places a significant burden on organisations regarding the handling of personal data. 
The perceived efforts and risks of complying with the GDPR further increase when data processing activities span across organisational boundaries, as is the case in both small-scale data sharing settings and in large-scale international data spaces.

This paper addresses these concerns by proposing a case-generic method for automated normative reasoning that establishes legal arguments for the lawfulness of data processing activities. 
The arguments are established on the basis of case-specific legal qualifications made by privacy experts, bringing the human in the loop.
The obtained expert system promotes transparency and accountability, remains adaptable to extended or altered interpretations of the GDPR, and integrates into novel or existing distributed data processing systems.

This result is achieved by defining a formal ontology and semantics for automated normative reasoning based on an analysis of the purpose-limitation principle of the GDPR.
The ontology and semantics are implemented in eFLINT, a domain-specific language for specifying and reasoning with norms.
The XACML architecture standard, applicable to both access and usage control, is extended, demonstrating how GDPR-based normative reasoning can integrate into (existing, distributed) systems for data processing.
The resulting system is designed and critically assessed in reference to requirements extracted from the GPDR. 
}




\maketitle

\newcommand{\art}[1]{Art. #1}
\newcommand{\artm}[2]{\art{#1}(#2)}
\newcommand{\artmm}[3]{\artm{#1}{#2}(#3)}

\section{Introduction}\label{sec1}

%The introduction of the GDPR privacy regulation in 2018 has had a significant impact on individual businesses and organisations as well as how businesses 

Organisations operating across a wide range of sectors are increasingly collaborating in the exchange of data and data-related services in order to leverage (shared) benefits out of the availability of data, such as improving efficiency in public governance, mobility or logistics, increasing the effectiveness of medicine, healthcare and science, or gaining insights into customer behaviour and market potential.
The value of data held by organisations is recognised at all levels, ranging from individual organisations, small-scale consortia, sector-level ecosystems, and the European Union (EU) striving for a single data market~\cite{eu-data-strategy}, as reflected in the Data Act and Data Governance Act regulations.
Simultaneously, the EU strives for data sovereignty for its citizens, as reflected in General Data Protection Regulation (GDPR) and the AI act.
The priorities of increasing the availability of data whilst retaining sovereignty are in tension, particularly when personal data is used within autonomous systems.
In this paper we aim to reduce this tension by proposing an approach for partially automated GDPR-reasoning within distributed data processing systems. %, intended to increase the efficiency and accountability of (lawfully) processing personal data.

To make the most out of data, data processing systems can ideally employ data and data-related services from many available sources and process data largely automatically and efficiently.
On the other hand, compliance with the GDPR requires careful consideration of various aspects regarding the use of personal data, such as the purpose of processing and data minimisation with respect to the processing purpose. 
The required care is exacerbated by the inherent complexity of the GDPR, comprised of many open texture terms (e.g., the `incompatibility' of purposes) and provisions that apply only to specific situations (e.g., relating to special categories of personal data).
Moreover, the lawfulness of processing activities needs to be reassessed periodically and the decision-making should be accountable, i.e., controllers are responsible to retain information that enables reporting to both data subjects and internal and external auditors.  
Thus, the legal requirements placed on a software system by the GDPR tend to be case-specific and demand constant attention by policy makers and privacy experts within an organisation. 
As a result, the data governance and privacy policies of organisations are typically manually converted into software requirements and code, resulting in data processing systems that need to be adapted (reprogrammed) when policies change, which is not only costly, but increases the risk of non-compliance and the difficulty of demonstrating compliance.
And especially in a context of cross-organisational data sharing, defensive policies may be put in place to mitigate perceived risks, hampering collaboration. 
\begin{figure}
    \centering
    \includegraphics[width=0.5\linewidth]{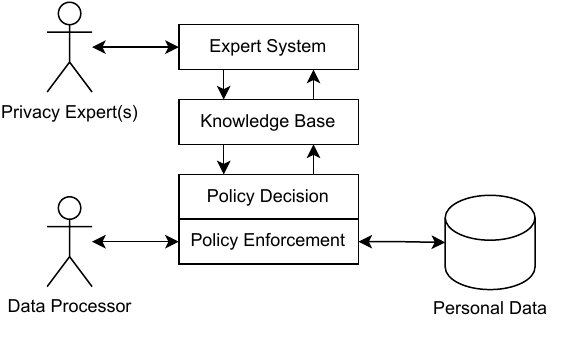}
    \caption{High-level diagrammatic representation of the system proposed in this paper. The system can be used in conjunction with existing policy enforcement mechanisms within a system.}
    \label{fig:expert-system}
\end{figure}
In this paper, for explainability and auditability reasons, we propose the use of a rule-based expert system (visualised in \Cref{fig:expert-system}) to reason about the lawfulness of data processing actions.
Case-generic parts of the GDPR, e.g., that `legal obligation' is a possible basis for lawful data processing, are interpreted and converted into rules within a domain-specific language (DSL) for normative reasoning.
Case-specific parts, e.g., which legal basis is used and whether data subjects have been informed about the purpose of processing, are provided as qualifications by one or more privacy experts interacting with the system.
Together, the rules and qualifications form a knowledge base for automated reasoning about the lawfulness of individual processing actions.
The outcome may be that the processing action is not lawful, requiring (and helping) the privacy expert(s) to reconsider their analysis of the case.
Alternatively, the outcome may be an argument that the processing action is lawful, in which case an authorisation for the action is generated within an access or usage control system that governs physical access to data.

The proposed approach has multiple advantages.
Firstly, by decoupling the (encoded) interpretation of case-generic parts of the GDPR, the overall system is inherently more transparent and adaptable in its integration of the GDPR: a modification or extension to the interpretation of the GDPR affects only the rules written in the DSL and leaves the rest of the data processing system intact.
Secondly, by demanding case-specific qualifications from users, the closure of open texture terms and other context-specific decisions are made only when the needed context information is available.
Thirdly, by requesting and explicating these qualifications, the system can aid in complying with the GDPR requirements of periodic review and accountability (by systematically recording yet periodically invalidating individual qualifications). 
As such, the proposed system occupies a position in the spectrum of partial automation in which crucial responsibilities are left to human actors whereas logical reasoning and administrative tasks are delegated to automated components of the system. 

In reference to \Cref{fig:expert-system}, this paper firstly focusses on establishing the knowledge base of rules and qualifications, forming the basis for the automated construction of lawfulness arguments and authorisation for processing actions.
Secondly, the paper describes different ways in which the automated reasoning procedure can integrate in distributed system, building on top of the distributed architecture of the XACML access control standard. 
In doing so, the application of normative reasoning to both ex-ante and ex-post scenarios is discussed.
The paper does not, however, discuss the expert system and its possibly many interactions with privacy expert(s).
For example, the expert system may assist the expert in establishing a legal argument, in communicating with data subjects, and in generating reports for auditing purposes.
Matters of human-computer interaction, user experience and explainability are left as future work. 
And rather than providing a complete analysis of the GDPR, the paper instead demonstrates the concepts laid out above by considering the purpose-limitation principle and GDPR-provisions regarding the lawfulness of processing.

The paper makes the following contributions:
\begin{itemize}
    \item An analysis of the purpose-limitation principle of the GDPR results in a set of requirements and an interpretation of norms formalised as an ontology with associated semantics given as a system of logical inference. The result is case-generic and transfers to other application contexts. The semantics can be operationalised in general-purpose languages, logic programming languages and database languages. The eFLINT DSL is used for an example implementation.
    \item Varying approaches to integrating the resulting normative reasoning procedure within distributed data processing systems are discussed. Variants show different patterns of delegating responsibilities between controllers, processors and governance intermediaries. By relying on the well-established XACML standard, integration can be realised in a wide range of systems, including systems with existing access and usage control implementations. 
    \item The distribution of responsibilities between human and software actors helps organisations achieve compliance in an accountable and efficient manner. 
\end{itemize}

The paper is outlined as follows.
\Cref{sec:background} gives relevant (technical) background information and related work. 
\Cref{sec:purpose-limitation} analyses the lawfulness of data processing according to the GDPR, focussing on the purpose limitation principle. 
The analysis results in a set of requirements and the definition of an ontology in \Cref{sec:ontology}.
The ontology is given a formal semantics in \Cref{sec:semantics} and an example implementation and application \Cref{sec:implementation}.
\Cref{sec:integration} presents the integration of GDPR-based normative reasoning in distributed data processing systems, focussing on ex-ante scenarios in which authorisation is produced for individual processing actions.
\Cref{sec:ex-post-scenarios} reflects on the application of the normative reasoning procedure in ex-post scenarios such as internal and external audits.
\Cref{sec:discussion} evaluates the extent to which the requirements have been met and critically reflects on the advantages and limitations of the approach.
\Cref{sec:conclusion} concludes.
\section{Technical Preliminaries and Related Work}
\label{sec:background}
\label{sec:preliminaries}
This section gives a brief summary of related work and technical preliminaries, introducing concepts and terminology applied throughout the rest of the paper.
The following topics are discussed: distributed data processing systems, (purpose-based) access and usage control, logical inference, and the normative specification language eFLINT~(\cite{eflint}), used in our implementation.

\subsection{Distributed Data Processing and Data Exchange Systems}
Organisations are expanding their collaborations with other organisations in attempts to extract value and insights out of available data.
For example, organisations collaborate to increase the amount of data available for training a machine learning application, or for finding certain patterns.
These collaborations can be as simple as one-to-one \emph{data sharing} -- making data available to another organisation, perhaps after some filtering, transformation and/or anonymisation. 
More complex collaborations involve the execution of entire data science workflows, in which both data and algorithms are potentially delivered by separate organisations.
We refer to systems that support such complex collaborations as \emph{data exchange systems}.

The first type of collaboration are sometimes supported by `data marketplaces' in which datasets (and other data-related assets) are offered between organisations~(\cite{Spiekermann2019}). 
A data marketplace typically offers publishing, finding and acquisition services, optionally including the handling of payments and logging transactions.

The more complex collaborations require services for scheduling and orchestrating workflow execution, resulting in processing actions distributed across nodes in a computer network~(\cite{Cohen2014}).
In the context of computational science, e-Science and GRID-computing, distributed workflow execution is commonplace and can be optimised for various criteria, such as runtime performance~(\cite{Degeler2014}), bandwidth~(\cite{Saewong1999}) or reduced energy consumption~(\cite{CHEN2023102938}).
Various authors have proposed data marketplaces supporting workflow execution ~(\cite{zhang2019,shakeri2020,mahiru,Kassem2021}).
Research into the various kinds of data exchange systems described here is executed in the context of the International Data Spaces Association (IDSA) (see, e.g., \cite{idsa-ram, Jung2022}) and the Amsterdam Data Exchange (AMdEX) (see, e.g., \cite{amdex-ra}). 
In both initiatives, usage control is a key element to achieve control and trust (or reduction of risk) to the stakeholders involved.

In data exchange systems, it is important to consider that the processing activities can be distributed across organisational boundaries.
Such distribution is relevant from the legal perspective as different organisations may reside in different jurisdictions and may have different legal positions.
For the purposes of this paper, we presume all mentioned organisations and actors reside in the EU and are consequently subjected to the GDPR.
A key topic of discussion in this context is whether an organisation is considered a controller or a processor according to the GDPR when assigning responsibilities within the data exchange system.
Another connected key topic, and the focus of this paper, is the discussion on the lawfulness of the individual processing activities.

Within a data exchange system, organisations can have different responsibilities, both from the legal perspective (e.g., controller or processor) and the technical perspective (e.g., storing, transforming or receiving data). 
If the assignments of legal and technical responsibilities diverge, then the assignment is likely to cause non-compliant behaviour. 
In addition to the GDPR, organisations may have their own policies regarding privacy, security and data governance. 
In general, it is difficult to align the interests of all organisations collaborating with a data exchange system.
These difficulties are exacerbated by the need to also consider additional stakeholders such as data subjects and relevant authorities.  
The research presented in this paper is part of a wider research agenda addressing the challenges laid out above.

Access and usage control models, discussed in the next subsection, describe methods, of varying strengths, to deliver control to individuals by setting \emph{policies}.
The policies are administered separately, realising an important separation of concerns: access and usage policies can evolve independently from the system architecture and its implementation.
One of the contributions of this paper is raising the level of abstraction of (at least part of) the policies such that they can be given by a domain-expert, i.e., a privacy expert such as a privacy officer.

\subsection{Access and Usage Control}
\label{sec:background-xacml}
\begin{figure}
    \centering
    \includegraphics[height=2.2cm]{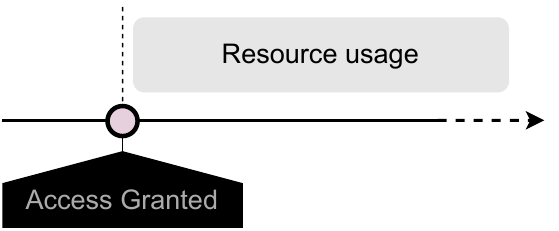}
    \caption{The enforcement process of an access control system. There is no further control after access has been granted and the resource is used.}
    \label{fig:time_access_control}
\end{figure}
%
% This subsection discusses existing access control and usage control models and their application within the context of data processing.
%
% We consider two dimensions of distribution: (1) the distribution of processing actions over time and (2) the distribution of processing actions across system nodes and organisational boundaries. 
%

\textbf{Access control} is a popular means to protect digital resources from unwanted processing actions performed on those resources. 
These resources may be dataset, algorithms, or more fine-grained data elements such as specific database tables and attributes.
Similarly, the processing actions may vary in level of abstraction, including ``read'' and ``write'' actions within a database system or ``play'' and ``distribute'' within a digital right management (DRM) system. 
Access is by default restricted, and must be explicitly granted through policy rules that determine whether a particular actor can perform a specified action on the resource, in which case authorisation to perform the action is provided.
In what follows, we use the term `authorisation' to refer to the technical term related to access control, whereas `permission' refers to the legal concept.
In the ideal situation, authorisations and permission coincide, i.e. there are no authorisations without permissions and vice versa.

In a conventional access control models -- such as Role-based Access Control (RBAC) as in~\cite{rbac} or Attribute-based Access Control (ABAC) as in~\cite{abac} -- an access request identifies the actor making the request, the resource on which the action is to be performed and the (type of) action to be performed (e.g., read or write).
Policy rules distinguish access requests based on these elements, and consider additional attributes such as the `role' the actor plays within an organisation, the security clearance of the actor, and/or the confidentiality level of the resource. 
These simple and powerful models cover many real-world scenarios and are widespread in, for example, the health-care domain~\cite{}.
Frequently applied standards are the eXtensible Access Control Markup Language (XACML)\footnote{\url{https://www.oasis-open.org/standard/xacmlv3-0/}} and the Open Digital Rights Language (ODRL)\footnote{\url{https://www.w3.org/TR/odrl-model/}}.

Through the use of attributes in ABAC models, resources and actors can be associated with different organisations and policies can use this information to grant or restrict access~(\cite{abac}). 
Such access control models can therefore distinguish between processing actions distributed across actors, organisations and system nodes.
However, conventional access control models do not consider processing actions as distributed across time.
An access control system only enforces the policies once, directly prior to the usage of the resource, corresponding to \emph{ex-ante enforcement} of the policies.
As a result, a system administrator cannot control, through the expression of policies, what happens with the resource after access has been granted and the resource is being used.
More powerful control mechanisms are needed to ensure that data is, for example, not further distributed, if the `read' action is permitted but the `distribute' action is not.
\Cref{fig:time_access_control} lays out the events related to an access control request over time.

\begin{figure}
    \centering
    \includegraphics[height=3.5cm]{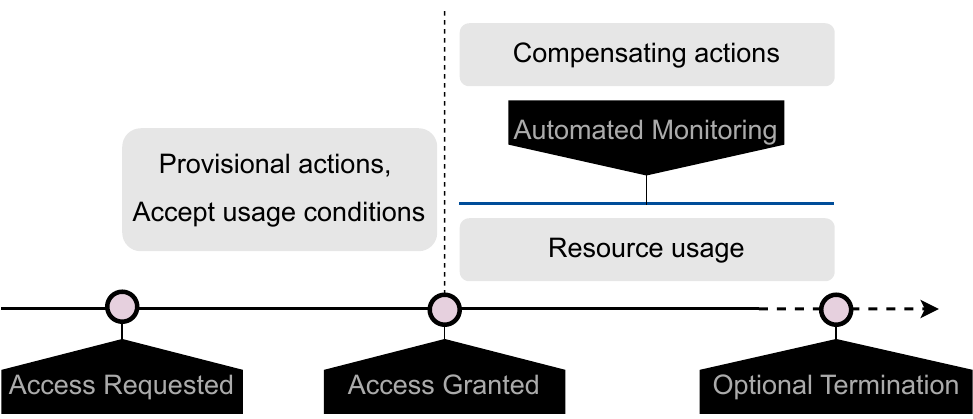}
    \caption{The enforcement process of a usage control system with continuous monitoring and control through compensating actions. The diagram is an adaptation of Figure~3a in~\cite{Pretschner2006}.}
    \label{fig:time_duc}
\end{figure}
\textbf{Usage control} is a generalisation of access control that considers the time dimension in several regards.
As introduced by Sandhu et al., usage control extends access control with the notion of ``continuity'' and ``mutability''~(\cite{10.1007/978-3-540-45215-7_2}).
\emph{Continuity} refers to the on-going evaluation and enforcement of authorisations throughout the usage of a resource.
The UCON+ model of~\cite{Hariri2023} extends the notion of continuity of the original UCON model with \emph{continuous monitoring} to trigger re-evaluation of authorisations when changes in attributes (e.g. of the actor, resource or the wider environment) are observed. 
A system with this form of continuity can thus realise dynamic forms of ex-post enforcement in which violations are observed and authorisations can be revoked and processing terminated\footnote{Although more powerful penalising powers may be needed for ex-post enforcement of legal obligations, in particular, as discussed by~\cite{10.1007/978-3-030-89811-3_4,chowdhury2012}.}. 
\emph{Mutability} refers to the ability for the attributes on which access decisions are made to be modified as a result of the execution of actions by actors.
A system with mutability can thus realise the legal concept of power, assuming attributes play a role in determining the legal positions of actors.
\Cref{fig:time_duc} lays out the events related to a usage control request over time. 
Different methods and techniques for implementing usage control exist, but an elaborate description is outdside the scope of this paper.

The purpose for processing the resources is not being addressed in usage control systems, while purpose is a pivotal concept with the GDPR. The purpose-based access control model presented in this paper is an extension to such usage and access control models and could in fact be built upon existing usage and access control systems. 

%does not depend on the more powerful form of control offered by usage control models compared to access control models.
%
%As a result, our approach can be applied both in the context of (existing) access control and usage control implementations.
%
In \Cref{sec:architecture} we give an example of how one could realise purpose-based access control using the XACML architecture for access control systems as a basis. 

\textbf{Purpose-Based Access Control} (PBAC) models extend access control models by explicitly introducing the notion of purpose into the process of producing authorisations.
Diverse methods for specifying and implementing purposes within access control systems have been introduced, briefly summarised in what follows. 

Early work on PBAC, such as \cite{byun2005purpose}, defines purpose as the underlying rationale for data collection and access. 
In their work, purposes are organized hierarchically, and users are authorized for specific access purposes, allowing them to access data for those purposes. 
\cite{zhang2019purpose} introduce a purpose graph representing all possible access purposes in a system, allowing purposes to have multiple ancestors. 
\cite{pallas2020towards} propose a purpose-based access control framework at the application layer, establishing a purpose hierarchy based on compatibility relationships defined in a configuration file. Queries are enhanced with purpose-related attributes.

Other approaches focus on purpose specification using privacy policy languages. 
For example, the Enterprise Privacy authorisation Language (EPAL) that is designed to express enterprise privacy policies governing data management~(\cite{ashley2003enterprise}). 
EPAL represents purposes as atomic values that model intended data processing services. 
While EPAL determines purposes before policy evaluation, it does not directly validate them, relying on the application for validation. 
In the Purpose-to-Use (P2U) language introduced by \cite{iyilade2014p2u}, purpose policies specify data sharing purposes, recipients, and retention periods.

The PAL language by~\cite{10.1007/978-3-031-57978-3_4} integrates GDPR roles, purpose and consent directly into a general-purpose programming language based on the Active Object programming model.
A formal semantics for the language is used to argue that all executions of programs written in the language are compliant according to the GDPR provisions considered.
The analysis of the GDPR in this paper is limited to consent as a legal basis and does not consider other legal bases nor specificity and compatibility of purposes.
The ontology and semantics we consider in this paper can potentially be used to extend the syntax and semantics of PAL.
%
%Realising such extensions is not a simple task however, as the GDPR provisions are encoded directly in the semantics, and may require breaking changes.
%
Note that in our approach, the precise interpretation given to the GDPR provisions is decoupled from the program being controlled, increasing maintainability, transparency and accountability. 

\cite{de2015declarative} present a declarative framework grounded in first-order temporal logic which offers precise semantics of purpose in policies. 
Purpose is defined as actions within a plan to achieve that purpose, expressed using temporal logic. 
\cite{jafari2009enforcing} explore purpose enforcement in workflow-based access control. 
An `access purpose' aligns with the intended `resource purpose', organized into abstract and concrete purposes. 
The work by \cite{tschantz2012formalizing} delves into purpose semantics, using a formalism based on planning and modified Markov decision processes to determine if action sequences align with specific purposes. 
Lastly, \cite{basin2018purpose} and \cite{petkovic2011purpose} equate purposes with business processes and demonstrate the role of formal inter-process communication models in auditing and deriving privacy policies.

\subsection{Distributed Policies and Control}
\begin{figure}
    \centering
    \includegraphics[width=0.45\linewidth]{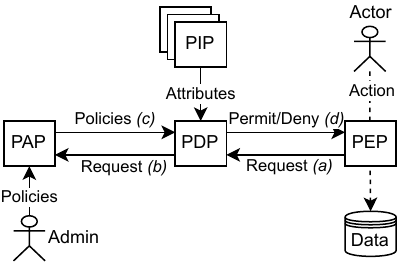}
    \caption{A simplified version of the XACML architecture. Compared to version 3.0 of the standard, the context handler component has been omitted.}
    \label{fig:xacml-roles}
\end{figure}
The integration of access or usage control in a \textit{distributed} system, such as a data exchange system, raises the question in what ways the steps of an access/usage control process can be distributed across system components, and which distribution is preferred in different contexts.
The technical roles captured by the XACML architecture, of which a simplified version is presented in \Cref{fig:xacml-roles}, enable this discussion, which we leverage in \Cref{sec:integration}.

In this high-level architecture, an administrator determines which policies are applicable to (upcoming) requests by registering policies in the policy administration point (PAP). 
%
% An actor attempts to perform an action on a data-source.
%
The policy enforcement point (PEP) is a control mechanism that conditionally prevents the execution of an action by an actor on a (data) asset.
The action is admitted only if the policy decision point (PDP) permits this specific access request, identified by the actor, action, and asset (as described earlier). 
The PEP forwards the request to the PDP and awaits its permit or deny decision.
The PDP receives the set of relevant policies by forwarding the request to the PAP and uses these policies, together with additional policy information provided by policy information points (PIPs), to make its decision.
Policies administered in the PAP are associated with `targets' determining which policies are applicable to which requests.
The PIPs provide additional policy information, such as roles and attributes (as described earlier).
Which policy information is available is dependent on the specific access control model applied and the ability to monitor or extract the required information from within the system.

These technical roles (PDP, PAP, PEP, and PIP) are useful to describe (patterns of) system architectures at a high conceptual level, a feature taken advantage of in \Cref{sec:integration}.
It is important to note that some of the technical roles may be assigned to a single component in the architecture of a system, depending on the level of abstraction with which the system is described or the desired distribution of responsibilities.

\subsection{Logical Inference, Logic Programming and Explainability}
\label{sec:background-inference}
\newcommand{\ruleseplength}{1em}
An important aspect of our contribution is the assignment of a formal semantics to elements of the PBAC model we propose.
The formal semantics, given in \Cref{sec:semantics}, is in the form of an inference system based on first-order logic. 
The inference system defines logical predicates, closely aligned with elements of the ontology of our model.
The inference system consists of a number of inference rules that are used to infer the truth-values of the predicates.
Every inference rule is of the following form:
\begin{equation}
\inferrule{
\mathit{antecedents}
}{
\mathit{conclusion}
}
\end{equation}
An inference rule can be read as a logical implication from top to bottom, i.e., if all antecedents hold, then the conclusion follows.
The conclusion and antecedents are {terms} referring to predicates and possibly involving placeholders.
For example, the following inference rules define the predicate $\mathit{sibling}$ in terms of the predicates $\mathit{brother}$ and $\mathit{sister}$, respectively, and contain the placeholders $A$ and $B$.
\begin{gather*}
    \inferrule{
    \mathit{sister}(A, B)
    }{
    \mathit{sibling}(A, B)
    }[\textsc{sister}]\quad
    \inferrule{
    \mathit{brother}(A, B)
    }{
    \mathit{sibling}(A, B)
    }[\textsc{brother}]
\end{gather*}
Assuming \texttt{alice}, \texttt{bob}, and \texttt{chloe} are valid elements of the domains of the mentioned relations, one can ask the question whether \texttt{bob} and \texttt{chloe} are siblings, represented by the predicate $\mathit{sibling}(\texttt{bob}, \texttt{chloe})$.
One can infer the truth of this statement through applications of the inference rules as follows: for each of the inference rules with a conclusion referring to $\mathit{sibling}$, substitute the placeholders in the rule such that the conclusion reads $\mathit{sibling}(\texttt{bob}, \texttt{chloe})$ (if possible) and attempt to infer the resulting antecedents.
If this recursive process succeeds for at least one of the inference rules, an inference has been found.
Thus, where there is a logical conjunction between the antecedents (all have to be inferred), there is a logical disjunction between the rules (only one has to be applicable).

Applying this process to the example question and rules produces the following concrete instances of the rules:
\begin{gather*}
    \inferrule{
    \mathit{sister}(\texttt{bob}, \texttt{chloe})
    }{
    \mathit{sibling}(\texttt{bob}, \texttt{chloe})
    }\quad
    \inferrule{
    \mathit{brother}(\texttt{bob}, \texttt{chloe})
    }{
    \mathit{sibling}(\texttt{bob}, \texttt{chloe})
    }
\end{gather*}
The top elements of these concrete rules are referred to as `goals' that require further inference.
For the process to complete, inference rules without antecedents, referred to as \emph{axioms}, are required.

Consider the following rules, of which the rule on the right-hand side is an axiom:
\begin{gather*}
    \inferrule{
    \mathit{sister}(B, A)
    }{
    \mathit{brother}(A, B)
    }[\textsc{symmetry}]\quad
    \inferrule{
    }{
    \mathit{sister}(\texttt{chloe}, \texttt{bob})
    }[\textsc{axiom}_1]
\end{gather*}
With the given set of rules, one can now infer that, indeed, $\mathit{sibling}(\texttt{bob}, \texttt{chloe})$.
This is achieved through applications of the rules \textsc{brother}, \textsc{symmetry}, and finally $\textsc{axiom}_1$. 

The result of an inference process, referred to as a \emph{derivation}, can be visualised as a tree.
The derivation (tree) of the previous example is as follows:
\begin{gather*}
\inferrule{
  \inferrule{
    \inferrule{
    }{
    \mathit{sister}(\texttt{chloe}, \texttt{bob})
    }[\textsc{axiom}_1]
  }{
    \mathit{brother}(\texttt{bob}, \texttt{chloe})
  }[\textsc{symmetry}]
}{
\mathit{sibling}(\texttt{bob}, \texttt{chloe})
}[\textsc{brother}]
\end{gather*}
The derivation is complete as there is no remaining goal at the top. 
Note that each inference step is labelled with the name of the rule applied at that step.
Further note that branching will occur in the derivation tree when one or more of the applied rules has more than one antecedent, each resulting in a goal that requires inference.

Most relevant to this paper is that derivation trees not only provide evidence to the successful completion of inference, but also show which rules have been applied.
This makes it possible to determine cause and effect, i.e., a particular conclusion was drawn, \emph{because} certain rules were applicable and certain axioms held true. 
This important quality of inference systems and logic programming languages has been linked to the concept of explainability.
As shown by, for example, \cite{Satoh2023} and \cite{Kowalski2023}, derivation trees can be post-processed to produce alternative representations such as more human-readable diagrams and natural language texts.
Depending on the application area, the generated representation can be targeted at different audiences, e.g., displaying a derivation tree as a legal argument for a legal expert. 
The innate ability of inference systems can be enhanced by associating documentation to axioms and inference rules such that the documentation can be incorporated into the explanations generated from derivation rules.
This possibility is an important motivation of our use of logic programming, although a demonstration is left as future work.

The relation between inference systems and logic programming languages such as Prolog~(\cite{Warren2023}) and Datalog~(\cite{datalog-book}) has been studied extensively.
A comprehensive overview has been given in the handbook by \cite{10.1093/oso/9780198537465.001.0001}.
These investigations are typically into the nature of the logic applied and the particulars of the inference algorithm. 
In \Cref{sec:semantics}, first order logic is applied and no specific assumptions on the inference algorithm are made.

Describing an inference system in a logic programming language necessarily involves an operationalisation determined precisely by the formal semantics of the chosen language.
This topic is left out of scope for this paper.
The implementation of the semantics in \Cref{sec:semantics}, given in \Cref{sec:implementation}, uses the eFLINT language.
The next section informally introduces the main constructs of eFLINT; 
the reader is referred to~\cite{eflint} and~\cite{VANBINSBERGEN2022140} for more details on the design of the language.
By separating semantic description from operationalisation, we highlight the fact that our approach does not depend on eFLINT and can be implemented in multiple languages.

\subsection{Normative Reasoning with eFLINT}
\label{sec:background-eflint}
The eFLINT language is a domain-specific language designed for encoding interpretations of laws, regulations, contracts and other normative documents~(\cite{eflint}). 
The legal foundations are provided by the framework of legal concepts by~\cite{hohfeld1917fundamental}.
The language is oriented towards actions: actions may be permitted or obliged and can have effects.
An act\footnote{An act, or action-type, describes a class of actions, i.e. an action is an instance of an act.} describes a \emph{power} when it has the effect of modifying the legal positions of one or more actors.
The obligation to perform an act is referred to as aa \emph{duty}.
The language has been explicitly designed to enable connecting physical actions to institutional actions, connecting the interpretation of various normative sources, and reasoning about the (action- and duty-)compliance of actors in scenarios.
The language is modular in the sense that an eFLINT program can be composed out of small fragments that might themselves represent compositions~(\cite{VANBINSBERGEN2022140}). 

An eFLINT program consists of two main parts: a \emph{normative specification} and a \emph{scenario} description.
A normative specification consists of fact-type, duty-type, event-type and action-type declarations. 
The type declarations define predicates (over atoms, sets and relations), as is common in logic programming. 
Derivation rules can be associated with types that are semantically similar to rules of the Datalog language~(\cite{datalog}) and can be read as implications or inference rules.
The following eFLINT example defines the predicates `sister', `brother` and `sibling' of the previous section, alongside the \textsc{symmetry}, \textsc{brother}, and \textsc{sister} rules.
\begin{listing}
\begin{lstlisting}[caption={An eFLINT specification defining a simple ontology with some derivation rules.},firstline=1,lastline=7]
Fact person  Identified by Alice, Bob, Chloe // type has three instances (atoms)
Fact sister  Identified by person1 * person2 // binary relation over persons
Fact brother Identified by person1 * person2
Fact sibling Identified by person1 * person2
     Holds when sister(person1, person2)     // rule SISTER
               ,brother(person1, person2)    // rule BROTHER
Extend Fact brother Holds when sister(person2, person1)  // rule SYMMETRY  
\end{lstlisting}
\end{listing}

In the listing above, rules \textsc{sister} and \textsc{brother} are part of the declaration of the fact-type \lstinline-sibling-. 
Rule \textsc{symmetry} has been introduced separately as an extension declaration for the fact-type \lstinline-brother-. 
The extension mechanism makes it possible to add rules independently in a declarative, incremental fashion.

The following listing shows a simple action-type and duty-type declaration.
%.
\begin{listing}
\begin{lstlisting}[caption={A simple action-type and duty-type declaration.},firstline=9,lastline=12]
Act ask-for-help Actor person1 Recipient person2
  Holds when sibling(person1, person2)
  Creates duty-to-help(person2, person1)
Duty duty-to-help Holder person1 Claimant person2
\end{lstlisting}
\end{listing}
The derivation rule associated with the action-type determines which instances of the action-type hold true and are therefore permitted.
As such, the derivation rule captures the norm ``a person can ask their sibling for help, creating the duty to help for the sibling.''
The action \lstinline+ask-for-help(Alice, Bob)+ thus denotes a power, as performing the action results in a duty for Bob.
A duty is commonly associated with one or more violation conditions as well as so-called `terminating acts' that have the effect of removing (satisfying) the duty. 
A further explanation of duties is not needed, however, as they do not play a role in the implementation of \Cref{sec:implementation}.
%

%TODO: is this paragraph needed?
In the example above, the norm is formalized using a \lstinline-Holds when- clause.
A \lstinline-Conditied by- clause could have been used in the same fashion.
The difference between the two is that \lstinline-Holds when- clauses are disjunctive, i.e., only one of the rules of the clause needs to hold for the predicate to hold, whereas \lstinline-Conditioned by- clauses are conjunctive, i.e. all rules of the clause need to hold for the predicate to hold.

The type declarations of an eFLINT specification determine a space of possible knowledge bases linking predicates to truth-values.
The \lstinline-Creates- clauses (and similarly for the \lstinline-Terminates- and \lstinline-Obfuscate- clauses that are not shown here) determine how knowledge bases can be modified through the execution of actions and events.
In general, action-type and event-type declarations define transitions between knowledge bases, thereby generating a transition system in which paths represent sequences of performed actions and occurred events.

The \emph{scenario} of an eFLINT program is a sequence of event-triggers and action-triggers that effectively describes a path in the transition system generated by the specification of the program.
A scenario can be seen as an imperative part of the program as action- and event-triggers effectively mutate state, i.e., the `current' knowledge base, by modifying the truth-values of predicates.
Actions that do not hold true at the time they are triggered raise violations.
Similarly, when a knowledge bases is encountered in which a duty is violated, this violation is raised.
A scenario can also contain Boolean queries that can be used to test the contents of current knowledge base.
So-called instance queries can be used to generate all the instances subject to some (optional) filtering criteria, resembling a selection in relational algebra and queries in database languages.
The following listing gives an example:
\begin{listing}
\begin{lstlisting}[caption={An eFLINT scenario demonstrating event- and action-triggers and two types of queries.},deletekeywords={to},firstline={16},lastline={25}]
Fact person  Identified by Alice, Bob, Chloe // type has three instances (atoms)
Fact sister  Identified by person1 * person2 // binary relation over persons
Fact brother Identified by person1 * person2
Fact sibling Identified by person1 * person2
     Holds when sister(person1, person2)     // rule SISTER
               ,brother(person1, person2)    // rule BROTHER
Extend Fact brother Holds when sister(person2, person1)  // rule SYMMETRY  

Act ask-for-help Actor person1 Recipient person2
  Holds when sibling(person1, person2)
  Creates duty-to-help(person2, person1)
Duty duty-to-help Holder person1 Claimant person2

.

+sister(Chloe, Bob).              // event-trigger for adding a fact to the KB
                                  // various facts are derived, including:
                                  // sibling(Bob, Chloe), ask-for-help(Bob, Chloe)
?Holds(ask-for-help(Bob, Chloe)). // Boolean query succeeds
?Holds(ask-for-help(Bob, Alice)). // Boolean query fails
ask-for-help(Bob, Chloe).         // action triggered
                                  // the duty to help Bob is created for Chloe
?Holds(duty-to-help(Chloe, Bob)). // Boolean query succeeds
?-person When sibling(Bob,person).// query generating all siblings of Bob
                                  // yields: person(Chloe)
\end{lstlisting}
\end{listing}

The implementation in \Cref{sec:implementation} defines an action-type \lstinline-request- that corresponds to an access or usage control request for processing a data asset. 
The action-type will be associated with derivation rules that determine whether the processing action is considered lawful according to the GDPR, based on the inference system of \Cref{sec:semantics}.
A strength of using eFLINT is that the action-type \lstinline-request- can be extended with further conditions, on top of lawfulness according to the GDPR,  e.g., as determined by the policies of the owner of the asset.
In general, all features of the eFLINT language are available to augment the specification of lawfulness within this paper with additional norms encountered in, for example, policies or data sharing agreements.
The next section gives our interpretation of (personal) data processing under to the GDPR, including lawfulness, purpose limitation and accountability.
\section{Data processing under the GDPR}\label{sec2}
\label{sec:purpose-limitation}
\label{sec:gdpr}
This section gives an account of the most important provisions of the GDPR related to lawful data processing.
In passing, requirements for the system proposed in this work are identified and printed in boldface.

The General Data Protection Regulation (GDPR) aims to protect the fundamental rights of natural persons and in particular the right to protection of their personal data (\artm{1}{2}).
The GDPR applies to all processing of personal data of natural persons residing within the EU (\artm{2}{1} and \art{3}), whereas personal data refers to all information about an identified or (directly or indirectly) identifiable natural person, referred to as the data subject (\artm{4}{1}).
`Data processing' under the GDPR should be widely interpreted, referring to any (set of) operation(s) performed on (sets of) personal data, such as structuring, organising, storing, modifying, disclosing by transmission, pseudonymising, anonymising, erasing or destructing (\artm{4}{2}).

\paragraph{Original Purpose Specification}
To enable data subjects to take well-informed decisions about the processing of their data, the GDPR obliges data controllers - those determining the purposes of the processing of someone's personal data (\artm{4}{7}) - to ensure the processing is transparent, lawful, and fair (\artmm{5}{1}{a}).
An important aspect of the transparency and lawfulness requirements is the data processing principle of \textbf{purpose specification} (\artmm{5}{1}{b}), i.e., the controller must determine a specified, explicit and legitimate purpose for the collection of personal data.
This purpose needs to be \textbf{sufficiently specific} such that the data subject is able to judge what processing will take place (e.g., before giving their informed consent). 
Data subjects should be able to understand for what purposes their data are processed and to assess the risk of the processing.

Lawful processing of personal data needs to be based on one of the \textbf{legal bases} stated in \artmm{6}{1}{a-f}.
In short, these respectively concern: consent given by data subject(s) (a), performance of a contract (b), compliance with legal obligations (c), protecting vital interests of a natural person (d), public interest and public task (e) and legitimate controller interest (f).
%
%While we implemented a check for all legal bases in Art. 6(1), we focussed on Art. 6(1)(a) and Art. 6(1)(b) as the check for the underlying contract or consent can be easily formalised.
%
% Medical data is considered a special category of personal data under artikel 9(1) GDPR. The sensitivity of the data has to be considered when analyzing the compatibility of purposes under article 6(4) GDPR. As article 6(4) GDPR is not applicable to processing based on consent or a legal obligation, a clear distinction between the two different regimes needs to be made.
%
In case of the legal basis of consent, consent must be freely given, specific, informed and unambiguous (\artm{4}{11}).
Data controllers must, as part of their transparency obligations, appropriately {inform data subjects} about the purpose for which their personal data will be processed.
Consent can only be specified if the data processing purpose is {communicated in a concise, transparent, intelligible, and easily accessible form} by the controller (\artm{12}{1} and \art{13}).
The {burden of proof} that freely given, informed and unambiguous consent was given rests with the data controller (\artm{7}{1}).
Data subjects have the right to withdraw their consent at any time, whereby withdrawing consent must be as easy as giving consent (\artm{7}{3}).

In the reasoning approach presented in later sections, one or more legal bases can be claimed as the bases for lawful processing by controllers.
The responsibility to determine which legal basis applies, and on which grounds, rests with the controller, i.e., the reasoning procedures computes with legal bases as input and does not compute them as output.
Each legal basis requires its own analysis to establish exactly when it may apply, which is out of scope for this paper.
In the cases of \artmm{6}{1}{c} and \artmm{6}{1}{e}, more specific requirements may be provided by (the national laws of) EU member states (\artm{6}{2}), complicating this analysis.
For scientific research, a detailed analysis of legal bases and purpose specification can be found in~\cite{10.1093/jlb/lsae001}.

% TODO: 
% - Purposes need to be made explicit:
% All information about the specification and assessment of the purpose needs to be made explicit and needs to be communicated to the data subject (article 13/14 GDPR).

\paragraph{Purpose Limitation in Further Processing}\label{subsec2.1}
% This is necessary to determine whether the processing is compliant.
%
The specified purposes for processing need to be lawful and legitimate (\artmm{5}{1}{a,b}).
The assessment of lawfulness and legitimacy can change over time and needs to be \textbf{repeated periodically}.

Data cannot be further processed in a manner incompatible with the purposes for which it was originally collected (\artmm{5}{1}{b}) with the notable exceptions of certain archival, historical, statistical, or scientific purposes.
The controller needs to assess whether the further processing purpose is not incompatible with the originally specified purpose for processing.
The required \textbf{compatibility assessment} needs to be made on a case-by-case basis and involves establishing a clear understanding of the expectations of the data subjects.
In the assessment the controller needs to include the following elements (Art. 6(4)):
\begin{itemize}
    \item The relationship between the specified purpose and the further processing.
    \item The context of the collection of personal data and especially the relationship between the data subject and the controller.
    \item The type of data that is collected, such as special categories of personal data or data relating to criminal convictions or offences.
    \item The possible implications the further processing entails for the data subject.
    \item The applied safeguards, such as anonymisation and pseudonymisation.
\end{itemize}
%
%To assess the potential incompatibility of the new purpose with the original purpose, the controller needs to answer the question whether the permission provided for by law or informed consent is not in conflict with the new goal(s).
%
If the controller's compatibility assessment concludes that the new purpose is not incompatible with the original purpose, the controller can apply the same legal basis as was originally applied.
However, when the two purposes are incompatible, the controller needs to re-assess whether a new legal basis can be found for the further processing, which may involve requesting consent.
Art. 5(1)(b) states that further processing for historical, scientific or statistic research, is not considered to be incompatible.
Such further processing must be subjected to appropriate safeguards.
%
%This is especially relevant where the processing for the specified purpose(s) is not based on consent but on another legal basis.
%
%Whenever the processing is based on consent, the subject should always be able to withdraw their consent.
%
%Withdrawing consent should be as easy as giving consent and means that their data should be deleted as there is no other valid legal basis (Art. 7(3)).

\paragraph{Accountability and Transparency}
\artm{5}{2} states that the controller needs to be able to prove that all processing is compliant with all principles stated in \artm{5}{1}, meaning that the data processing, including the technical system used, must be configured in such a way that the controller is able to provide all necessary information to prove compliance.
This \textbf{accountability} requirement for the controller, requires from the controller the ability to demonstrate that the processing of data is done in a lawful, fair and transparent way.
%
%As explained earlier, one of the conditions for the data processing to be GDPR compliant is that it must be based on one of the lawful grounds laid down in Art. 6(1).
%
The controller should therefore be able to show that for every processing activity a valid legal basis applies and that the processing is necessary for the specified purpose.
The controller must also demonstrate that the processing is limited to the purposes for which the data was initially collected or that further processing of the data is not incompatible with the purpose for which the data was collected (as explained above).
A technical system proposed in later sections assists controllers in satisfying the accountability requirement by demanding and recording applied legal bases, specified purposes, and the results of compatibility assessments.

The requirement of \textbf{transparency} involves showing that all transparency requirements laid out in the GDPR are met. 
According to \artm{12}{1} a controller should inform the data subject in an intelligible and accessible manner.
What constitutes an intelligible and accessible manner is dependent on the knowledge and background of the data subject.
To correctly inform the data subject before the data processing, the controller must provide them with, i.a., general information about the purposes for which the processing takes place, the periods data are stored for or the criteria to determine this period, and the legal basis the processing is based on (Arts. 13 and 14). 
This information is usually communicated through a `privacy policy'. 
Whenever the data is not obtained from the data subject themselves (Art. 14), the controller must provide information about the source from which the personal data originate, and if applicable, whether it came from publicly accessible sources. 

Next to actively providing generic information about the data processing, the controller has to provide, upon a data subject's request to access their personal data, individualised information about their data (\artm{15}{1}).
Whenever a data subject requests access to their data, the controller must provide them at individualised level with complete information about the purposes that subject's personal data was processed for, the exact categories of the personal data concerned, the recipients of that subjects' personal data, the storage periods, the source of the data, or the existence of automated decision-making.
The controller must communicate this information within one month after receiving the request, in a concise, transparent, intelligible, and easily accessible form, using clear and plain language (Art. 12(1)).

When a controller intends to further process personal data for a different legal basis or purpose, the controller \textbf{must inform} the data subject(s) about the new purpose \textbf{prior to the processing} (\artm{13}{3} and \artm{14}{4}) and provide additional relevant information specified in \artm{13-14}{2}.

The technical system described in the next sections helps controllers satisfy certain aspects of the accountability requirement.
It should be stressed though that the system presented does not directly enforce the (pro-active) communication about the processing of personal data as is required for the transparency requirement.

\section{Purpose based usage control model}
\label{sec:ontology}
\label{sec:model}
\begin{figure}
    \centering
    \includegraphics[width=0.85\linewidth]{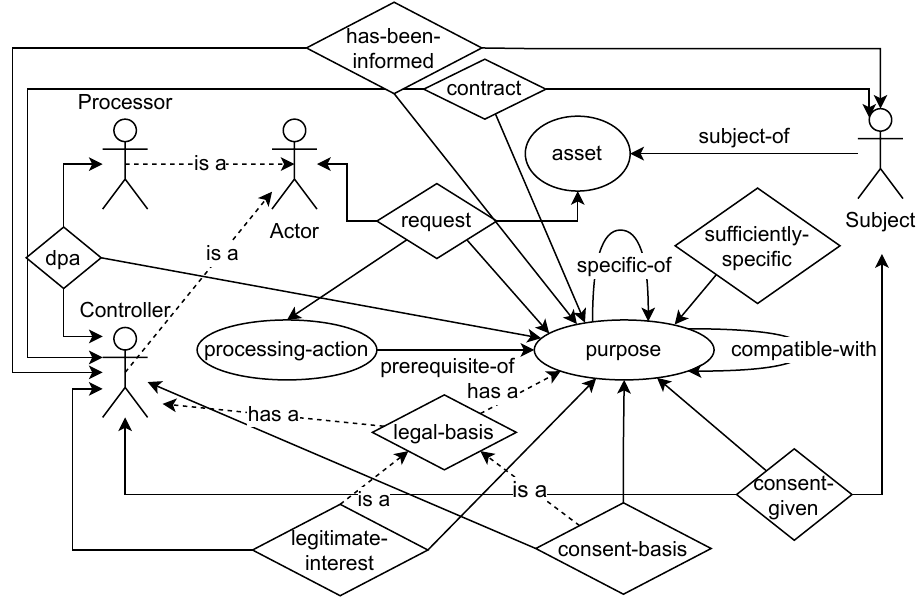}
    \caption{An ontology establishing relations between concepts as defined in \Cref{sec:purpose-limitation} and extended with a relation capturing processing requests. Concepts are represented as circles, relations as diamonds. Binary relations may be represented directly as a filled arrow between two concepts. The diagram shows only the legal basis relations derived from \artmm{6}{1}{a} and \artmm{6}{1}{f} for brevity.}
    \label{fig:ontology}
\end{figure}
This section defines the concepts and relationships between these concepts (an ontology, see \Cref{fig:ontology}) for usage control models based on the legal framework described in the previous section. 
The definitions that follow give a legal definition to the concepts identified in the ontology of \Cref{fig:ontology}. 
In \Cref{sec:semantics} a formal semantics is given to the same concepts, forming the basis for automated reasoning, as demonstrated alongside an implementation in \Cref{sec:implementation}.
The technical definitions are such that automated normative reasoning can derive authorizations from instances of the described concepts.
As such, this ontology is targeted to answer the essential question of purpose-based usage control: is the processing by an actor $a_i\in A$ (the set of actors) of an asset $d_j \in  D$ (the set of assets) for a stated purpose $p_k \in P$ (the set of purposes) legal according to the GDPR at a particular moment in time.
To answer this question, the normative reasoner needs to be provided with several elements of policy information.
%
% \Cref{fig:policy-roles} shows which actor roles contribute which elements of policy information to the reasoner.
%
For example, the controller determines the legal basis for a processing activity and data subjects contribute (or omit) their consent.
The roles and capabilities for providing policy information are discussed in \Cref{sec:architecture}.

\subsection{Definitions}
\label{sec:definitions}
Within the legal framework of the previous section a clear distinction between processing actions and purposes is made, resulting in distinct concepts in our ontology.
\begin{definition}
    A \emph{processing action} is the act of performing any operation on personal data (\artm{4}{2}).
\end{definition}
\begin{definition}
\label{def:processing-purpose}
    A \emph{(processing) purpose} is the goal for which the proposed processing of personal data is necessary.
\end{definition}
The processing purpose needs to be \textbf{sufficiently specific} for the data subject to be able to understand the nature of the processing and to give informed consent (see \Cref{sec:purpose-limitation}).
%

%
%This should be reflected by the access model to make sure that all processing is compliant with the GDPR.
%
The processing of personal data consists of actions, such as collecting, storing or transmitting, carried out by a processor on behalf of a controller or by the controller themselves.
%
% That is, both processors and controllers can 
%
Processing actions should always be carried out for a specific purpose for which the action is a prerequisite.
\begin{definition}
A \underline{processing action} is a \emph{prerequisite-of} of a \underline{purpose} when the processing action is necessary to fulfill the purpose. 
\end{definition}

A processing action operates on a data `asset'.
For reasoning about compliance, the contents of the data are irrelevant as long as it is possible to establish which data subjects are identifiable within the asset.
An asset is therefore be defined as a label that (uniquely) identifies an asset.
\begin{definition}
    An \textit{asset} is a label that refers to a dataset.
\end{definition}
In what follows, we may let the word `asset' simultaneously denote the label as well as the dataset to which the label refers where the ambiguity does not cause confusion.
The `subject-of' relation is introduced to capture the relation between assets and data subjects.
\begin{definition}
    A \underline{subject} is a \textit{subject-of} an \underline{asset} if the data of the subject is included in the dataset the asset refers to or the subject is otherwise identifiable within the data.
\end{definition}%
Note that we do not prescribe how the determination is made whether a subject is identifiable within the asset.

To ensure that the processing is not only bound to a specific purpose but that it is also lawful, the legal basis for the processing must be specified by the controller.
As discussed in \Cref{sec2}, processing must be based on one of the six legal bases in \artm{6}{1} of the GDPR~\cite{gdpr}.
In the ontology, `legal basis' is defined as an abstract relation between a controller and a purpose, with each of the members \artmm{6}{1}{a-f} contributing a more concrete version of the relation.
\begin{definition}
\label{def:legal-basis}
    A \underline{controller} can claim a \emph{legal basis} for processing for a specific \underline{lawful purpose} if the processing is lawful according to the GDPR (\art{6}), in which case one of the following applies (in summary):
    \begin{itemize}
        \item the data \underline{subject} has given consent (\artmm{6}{1}{a}), or
        
        \item the processing is necessary for: 
        \begin{itemize}
            \item the performance of a contract with the data, or
            \underline{subject} (\artmm{6}{1}{b}), or
        
            \item compliance with a legal obligation (\artmm{6}{1}{c}), or
        
            \item the vital interest of subject or natural person (\artmm{6}{1}{d}), or
        
            \item public interest or vested authority (\artmm{6}{1}{e}), or
        
            \item the controller has a legitimate interest (\artmm{6}{1}{f}).
        \end{itemize}
    \end{itemize}
    And all data subjects involved must be informed about the legal basis and purpose, prior to the processing.
\end{definition}
The ontology displayed in \Cref{fig:ontology} only shows the more concrete instances of the abstract `legal basis' relation only for the first and last clause of the definition above.
In our formalization, partly described in \Cref{sec:implementation}, all legal bases are included.
Note that in the cases of \artmm{6}{1}{a-b}, the relation is ternary rather than binary and involves a subject besides the required controller and purpose.

As we shall see in the technical definitions of \Cref{sec:implementation}, the legal basis relation forms the basis for generating authorizations.
In essence, authorization can be given for a processing action whenever it constitutes a prerequisite of a purpose for which a legal basis has been lawfully claimed by a controller.
The subject-of relation will be used to limit authorizations in cases where the legal basis for the processing requires this (i.e., consent or contract execution).
The requirement following from \art{6} can be made more flexible, however, generating authorizations also when the processing action is the prerequisite of a purpose that is more specific or not incompatible with the purpose of the claimed legal basis.
The following definitions describe the relations that capture these observations.

\begin{definition}
\label{def:specific-of}
    A \underline{purpose} is a \emph{specific-of} of another \underline{purpose} if it concretises a more abstract purpose without including elements not contained in the more abstract purpose. By definition, the specific-of relation is transitive, i.e., if \(p_1\) is a specific-of of \(p_2\) and \(p_2\) is a specific-of of \(p_3\), then \(p_1\) is a specific-of of \(p_3\). %
    % A legal basis is assumed by the specific-of relation as follows:
    % If \(p_2\) is a specific-of of \(p_1\), then \(p_2\) has a legal basis if \(p_1\) has a legal basis.
\end{definition}
Whether a stated purpose could be considered more specific than another stated purpose requires an, at least partially subjective, semantic assessment.
In our proposal, the controller makes explicit their determination of specificity, which can subsequently be assessed by a privacy authority.
This is similar to the assessment whether the processing purpose stated by the controller is sufficiently specific (see \Cref{sec:roles}).

As stated in \Cref{sec2}, processing can be lawful for another purpose whenever this purpose is not incompatible with the purpose of the original legal basis.
In our proposal, we introduce a `compatible with' relation between purposes that captures the active claim (by the controller) that the first purpose is not incompatible with the second.
\begin{definition}
A controller can claim a legal basis for further processing
\begin{itemize}
    \item if the (new) processing \underline{purpose} is \emph{not incompatible with} the \underline{purpose} of an existing legal basis, and
    \item if the controller has informed all involved data subjects about the new processing purpose, prior to the processing.
\end{itemize}
\end{definition}
The ontology (\Cref{fig:ontology}) contains a relation `compatible with' which is considered to be equivalent to `not incompatible with'.
However, note that \textit{in}compatibility may be easier to establish than compatibility.
Further note that `not incompatible with' can easily be derived from an incompatibility relation by enumerating the purposes that are not specified as incompatible.
An important design decision is thus whether to specify incompatibility or compatibility (equivalent to `not incompatible').
Choosing the first option may be more practical but implies that the absence of information, e.g., when forgetting to make an incompatibility assessment, can result in positive conclusions about lawfulness. 
The second option forces the assessment, as without the presence of a compatibility statement no positive conclusion about lawfulness on the basis of compatibility can be drawn (as formalised in~\Cref{sec:semantics}).
For reasons of safety we opt for the second option.
If desired, the controller can still apply a local derivation process to derive compatibility from the absence of incompatibility.
%
% In our view, a `compatible with' relation is preferred over an  `incompatible with' in which we check the absence of instances to establish whether two purposes are not incompatible.
%
% As shown in \Cref{sec:implementation}, our proposal forces the controller to make a compatibility-assessment in the cases where an assessment is needed, whereas in the alternative, forgetting to make an incompatibility-assessment may result in unlawful authorizations being generated by the model. 
%
%This relation is not symmetric as a big part of the compatibility analysis involves the expectations of the data subject.
%
% As the GDPR states that (further) processing should not be done if that is incompatible with the original purpose, we assume that it is the responsibility of the controller to make explicit that he considers two purposes compatible! This will result in defining this compatibility relation in the model.

The last relation in the ontology of \Cref{fig:ontology} is that of a processing request, modelled after the requests typically encountered in access and usage control models. 
In typical access and usage control models, a request is made by an actor, and specifies the action to be performed and the asset on which the action is performed.
In our proposal, the action is assumed to be a processing action on an asset containing personal data.
The actor is either the controller, or a processor performing the processing action on behalf of the controller.
The `request' relation is augmented with the \emph{processing purpose}, i.e. the goal for which the processing action is performed (see~\Cref{def:processing-purpose}).
The processing purpose of the request may not be identical to the (e.g. higher or lower abstraction level) purpose for which the data was originally collected and/or for which a legal basis has been established. 
In these cases, the processing purpose may still be lawful if the processing purpose is more specific than the original purpose or if the processing purpose is not incompatible with the original purpose, as captured by the following definition:
\begin{definition}
\label{def:request}
    A \emph{purpose-based processing request} connects an \underline{actor} (a processor or controller) to a processing \underline{action}, performed on an \underline{asset} for a prescribed \underline{processing purpose}. The request is considered lawful if:
    \begin{itemize}
   
        \item the action is prerequisite of the processing purpose, and
        \item the processing purpose is sufficiently specific, and
        \item the processing purpose:
        \begin{itemize}
        \item coincides with a purpose that has a lawful legal basis, or
        \item is more specific than a purpose that has a lawful legal basis, or
        \item is not incompatible with a purpose that has a lawful legal basis.
        \end{itemize}
    \end{itemize} 
\end{definition}
In practice, we expect the processing purpose to be determined by the controller directly if the controller performs the processing action.
Alternatively, if a processor is processing on behalf of the controller, the processing purpose is determined by the processor, that should coincide with the purpose described within the relevant data processing agreement (DPA, see \Cref{fig:ontology}) between the controller and the processor.
%
%In the semantics defined in the next section, the processing purpose is inferred from the \textit{prerequisite-of} relation. 
%
%Thus, by explicitly relating processing actions to purposes, one implicitly relates processing requests to purposes.
%
%This design decision is primarily made to ensure the contents of a processing request coincide perfectly with the contents of a typical access control request.
%

The next section defines a semantics for the ontology established in this section.
\section{Semantics}
\label{sec:semantics}
In this section we define a formal semantics using logical inference (see~\Cref{sec:background-inference}) for the concepts and relations of the ontology introduced in the previous section.
The concepts and relations of the ontology are defined as predicates.
For each predicate, its truth-values are determined either through stated facts or through applications of the logical inference rules presented in this section.
Facts are not provided in this section, but are instead provided by actors (according to their roles, as laid out in \Cref{sec:roles} and \Cref{tab:policy-roles}) in the form of (partial) purpose-graph contributions.
In what follows, the axioms are referred to as claims or qualifications, reflecting that facts may be disputed.

The logical inference rules define an inference system that captures the knowledge derived from the legal analysis regarding the lawfulness of data processing given in \Cref{sec:purpose-limitation}.
The inference system makes a formal connection between the lawfulness of the requested processing and an authorisation given on an individual request, denoted by the predicate \textit{lawful-request}.
In other words, if the predicate $\textit{lawful-request}$ holds for a particular processing request, then the processing is considered lawful and authorisation can be given. 
The process of inference determines whether the predicate holds for a particular request.
This is the essence of the normative reasoning for purpose-based control mechanism we propose in this paper.
%

% The result of inference is a derivation tree that demonstrates how lawfulness was established through an application of the inference rules.
% %
% And as has been done by~\cite{Kowalski2023}, derivation trees can be used to generate natural language text for the domain expert, e.g., in the form of a legal argument.
% %
% Moreover, inference processes can be automated, as is well understood in the context of logic programming, symbolic AI, and formal operational semantics.
% %
% In \Cref{sec:implementation} we automate inference through the use of eFLINT.
% %

In the rules, the placeholder $P$ is used for purposes, $A$ for actions, $U$ for generic actors, $C$ for controllers, $D$ for datasets, and $S$ for subjects.
Some auxiliary predicates are introduced to simplify the definition of the semantics.
Examples are the ternary relation $\textit{legal-basis}(P, C, D)$ that is specific to a dataset $D$ (unlike the abstract, binary \textit{legal-basis} relation of the ontology) and the relation $\textit{processor-for}(U, C, P)$ for determining whether the controller or a delegate processor $U$ performs the processing on behalf of the controller $C$ for purpose $P$.

The following inference rule establishes that processing is lawful if the processing action is a prerequisite action of a purpose for which a legal basis has been (rightfully) claimed\footnote{As discussed in \Cref{sec:roles}, the legal basis claim can be disputed, e.g., by a privacy authority.}.
Note that the predicate \textit{prerequisite-of} is established through qualification and is not defined through inference rules.
%
% \begin{equation}
% \label{eq:base}
% \small
% \inferrule{
% \textit{prerequisite-of}(A, P)\quad
% \textit{legal-basis}(C, P, D)\quad
% \textit{processor-for}(U, C, P) 
% }{
% \textit{request}(U, A, P, D)
% }
% \end{equation}
%
\begin{equation} 
\label{eq:base} 
\small 
\inferrule{\textit{request}(U, A, P, D) \quad \textit{prerequisite-of}(A, P) \quad \textit{legal-basis}(C, P, D)\quad \textit{processor-for}(U, C, P) }
{\textit{lawful-request}(U, A, P, D)}
\end{equation}

The actor $U$ should either be the controller, or a processor with a data processing agreement with the controller, stating the exact purpose for which the legal basis is claimed.
The following rules capture these semantics formally.
\begin{equation}
\small
\inferrule{
U = C
}{
\textit{processor-for}(U, C, P)
}\label{eq:controller-processor}
\end{equation}
\begin{equation} 
\small
\inferrule{
\textit{dpa}(C, U, P)
}{
\textit{processor-for}(U, C, P)
}\label{eq:processor-processor}
\end{equation}
The predicate $\textit{dpa}(C,U,P)$ captures the claim that a data processing agreement (DPA) exists between controller $C$ and processor $U$, and that it states purpose $P$.
%
% As a qualification, predicate $\textit{dpa}$ is not defined through inference rules.
%

As was established in \Cref{def:request}, a valid \textit{legal-basis} can be claimed by a controller in multiple ways, for which we use `legitimate interest' and `subject consent' as representative examples.
The following rules establish how a legal basis can be derived from these types of claims, which in the case of `subject consent' requires that consent has been given by all subjects identifiable in the dasaset (as determined by \textit{subject-of}).
\begin{equation}
\small
\inferrule{
\textit{legitimate-interest}(C,P)\quad
\textit{sufficiently-specific}(P)
\\\forall_S(\textit{subject-of}(S,D) \rightarrow \textit{has-been-informed}(S,C,P))
}{
\textit{legal-basis}(C, P, D)
}
\label{eq:legal-basis-consent}
\end{equation}

\begin{equation}
\small
\inferrule{
\textit{consent-basis}(C,P) \quad 
\forall_S (\textit{subject-of}(S,D) \rightarrow \textit{consent-given}(S, C, P))\quad
\textit{sufficiently-specific}(P)
\\\forall_S(\textit{subject-of}(S,D) \rightarrow \textit{has-been-informed}(S,C,P))
}{
\textit{legal-basis}(C, P, D) % note deviation wrt ontology
}
\label{eq:legal-basis-legitimate-interest}
\end{equation}
Similar rules for the other four types of legal bases are part of the inference system but have been omitted for brevity. 
The premises of these two rules are established as qualifications.
In all cases, the purpose stated in the claim needs to be (qualified as) sufficiently specific, as discussed in \Cref{sec:purpose-limitation}.
Also in all cases, all subjects need to have been informed of the purpose of the processing.
For some legal bases, it can be argued that subjects are automatically or implicitly informed.
For example, the execution of a contract between employer and employee may require the employer to process personal data of the employee for purposes that follow directly from the contract.
In these cases, the contract can be seen to inform the employee about data processing purposes.
To reduce effort of policy administrators, the following inference rule can be added to the inference system.
The rule lets a claim about the existence of a contract generate a claim about information given to the subject.
\begin{equation}
\inferrule{
\textit{has-been-informed}(S,C,P)
}{
\textit{contract}(S,C,P)
}
\end{equation}

The precise purpose for which a processing action is executed does not need to coincide exactly with the purpose for which a legal basis has been claimed.
As described in the previous sections, the processing action may be executed for a more specific purpose, or for a purpose that is not incompatible with the purpose for which a legal basis has been established. 
The following two rules capture these semantics (respectively).
%
% \begin{equation}
% \label{eq:specific-of}
% \small
% \inferrule{
% \textit{prerequisite-of}(A, P')\quad
% \textit{specific-of}(P', P)\quad
% \textit{legal-basis}(C, P, D) \quad
% \textit{processor-for}(U, C, P)
% }{
% \textit{request}(U, A, P, D)
% }
% \end{equation}
%
\begin{equation}
\label{eq:specific-of}
\small 
\inferrule{
\textit{request}(U, A, P, D) 
\quad\textit{prerequisite-of}(A, P)
\\\textit{specific-of}(P, P')
\quad \textit{legal-basis}(C, P', D)
\quad\textit{processor-for}(U, C, P')}
{\textit{lawful-request}(U, A, P, D)}
\end{equation}

% \begin{equation}
% \label{eq:compatible-with}
% \small
% \inferrule{
% \textit{prerequisite-of}(A, P')\quad
% \textit{compatible-with}(P', P)\quad
% \textit{sufficiently-specific}(P') \\
% \textit{legal-basis}(C, P, D) \quad
% \textit{processor-for}(U, C, P) 
% }{
% \textit{request}(U, A, P, D)
% } 
% \end{equation}
%
\begin{equation} 
\label{eq:compatible-with} 
\small 
\inferrule{ 
\textit{request}(U, A, P, D)
\quad \textit{prerequisite-of}(A, P)
\quad \textit{sufficiently-specific}(P)
\\\quad \textit{compatible-with}(P, P')
\quad\textit{legal-basis}(C, P', D)
\quad \textit{processor-for}(U, C, P')
\\\forall_S(\textit{subject-of}(S,D) \rightarrow \textit{has-been-informed}(S,C,P))}
{\textit{lawful-request}(U, A, P, D)}
\end{equation}

The explicit determination whether two purposes are not incompatible with each other, as captured by \textit{compatible-with}, is a qualification and has no inference rules.
The \textit{compatible-with} relation is not considered to be transitive (unlike \textit{specific-of}, below).
This design decision has primarily been made to force an expert-determination of (in)compatibility to gain authorisation, rather than partially computing the relation.
If lawfulness is claimed on the basis of compatibility, the processing purpose ($P'$, \Cref{eq:compatible-with}) is required to (also) be sufficiently specific.

The additional premise $\textit{sufficiently-specific}(P')$ is not needed in \Cref{eq:specific-of} as a purpose that is more specific than a sufficiently specific purpose should itself also be sufficiently specific, as captured by the following rule:
\begin{equation}
\inferrule{
\textit{specific-of}(P, P') \quad \textit{sufficiently-specific}(P')
}{
\textit{sufficiently-specific}(P)
}
\end{equation}

The \textit{specific-of} relation is given as a qualification. 
However, the predicate is also partly defined through inference rules.
In accordance with \Cref{def:specific-of}, the \textit{specific-of} relation is transitive, which is captured by the following inference rule.
\begin{equation}
\small
\inferrule{
\textit{specific-of}(P_1,P_2) \quad
\textit{specific-of}(P_2,P_3)
}{
\textit{specific-of}(P_1,P_3)
}
\end{equation}

For pragmatic reasons, \textit{specific-of} may also be considered reflexive, i.e., each purpose is then more specific than itself, as expressed by the following rule:
\begin{equation}
\small
\inferrule{
}{
\textit{specific-of}(P_1,P_1)
}
\end{equation}
The suggested reflexivity of \textit{specific-of} is practical as \Cref{eq:base} is then superseded by \Cref{eq:specific-of} (i.e., by taking $P = P'$).
Thus, \Cref{eq:base} can be ignored, simplifying the inference system and inference process.

\paragraph{A reflection on data usage requests}
As noted earlier, a typical access and usage control request consists of the triple actor~$U$, action~$A$, and asset~$D$, whereas the request predicate defined above has an additional processing purpose $P$.
A design decision can be made to include a ternary variant of the $\textit{lawful-request}$ predicate and an inference rule that selects the processing purpose based on a qualification that establishes a relation between processing actions and the processing purposes, separately from any requests.
The following rule defines the ternary variant of $\textit{lawful-request}$ based on the additional predicate $\textit{processing-purpose-of}(A, P)$ and the quaternary variant of $\textit{lawful-request}$:
\begin{equation}
\inferrule{
\textit{processing-purpose-for}(A, P) \quad \textit{lawful-request}(U,A,P,D)
}{
\textit{lawful-request}(U,A,D)
}
\label{eq:ternary-request}
\end{equation}
Adding \Cref{eq:ternary-request} to the inference system not only aligns the predicate with access and usage control requests, but it also adds the relation `processing purpose of' to the purpose-graph, which has practical implications.
In~\Cref{sec:architecture} it is suggested that the qualifications embedded in the purpose graph can be provided by privacy experts, whereas a data access request is typically performed by a professional fulfilling a different kind of role (e.g., a data scientist, doctor, or a member of human resources). 
These observations raise the following question: is the processing purpose to be determined (a priori or a posteriori) by a privacy expert, by the professional about to use data, or can this determination perhaps even be made automatically (e.g., by analysing the source code of an algorithm). 
This question is particularly relevant if the responsibilities of providing qualifications for the purpose graph and making data usage requests are spread over two (or more) organisations.
In what follows, we ignore \Cref{eq:ternary-request} and consider the professional that requests data usage responsible for providing the processing purpose (perhaps in consultation with a privacy expert of the controller). 

This concludes the definition of the semantics.
\Cref{sec:implementation} presents an implementation of the ontology and the semantics, and demonstrates normative reasoning in several scenarios.
\Cref{sec:architecture} discusses the roles and responsibilities involved when applying our purpose-based normative reasoning within data exchange systems.
\section{Implementation}
\label{sec:implementation}
This section implements the ontology and associated semantics described in the previous sections. 
The implementation is in the form of a specification written in the normative specification language eFLINT (\cite{eflint}).
A brief introduction to the language has been given in~\Cref{sec:background-eflint}.
The code is demonstrated with an example in \Cref{sec:impl-example}.
Some readers may find it beneficial to skip ahead and see the example application before the implementation.
An interpreter with which the eFLINT code of this section can be executed is available online\footnote{\url{https://gitlab.com/eflint/haskell-implementation}}. 
The code itself will be made available as supplementary material to this paper.

\subsection{Ontology and Semantics}
\label{sec:implementation-semantics}
The ontology (\Cref{sec:model}) and the predicates (\Cref{sec:semantics}) are implemented as type definitions. 
The inference rules of \Cref{sec:semantics} are implemented as derivation rules.

The following code fragment defines the actors of the ontology as fact-types.
The \lstinline-Derived from- derivation clause ensures that all instances of \lstinline-controller- and \lstinline-processor- also instantiate \lstinline-actor-, implementing the `is a' relationship as given in \Cref{fig:ontology}.
Subjects are not considered as actors and therefore do not make processing requests.
{
\begin{lstlisting}[caption={Specification of actors and GDPR roles.},label={listing:gdpr-actors},linerange=3-6]
Fact subject    
Fact controller 
Fact processor  
Fact actor Derived from controller, processor
\end{lstlisting}}

The following fragment defines the remaining relations and the `subject of' relation.
{
\begin{lstlisting}[caption={Specification of core ontology concepts.},label={listing:gdpr-concepts},linerange=9-13]
Fact processing-action 
Fact purpose        
Fact asset          
Fact subject-of Identified by subject * asset
Extend Fact subject Derived from subject-of.subject
\end{lstlisting}}\noindent%
The last line generates instances of \lstinline-subject-, deriving these instances from the first element of the `subject of' relation.
This rule is added to ensure that when the \lstinline+subject-of+ relation is given as a qualification, then the set \lstinline-subject- is derived from it.
The rule is defined using the \lstinline-Extend- keyword for adding additional clauses to an already existing type-definition.

With these definitions in place it is possible to define types for the \textit{request} and \textit{lawful-request} predicates, as is done in the fragment below.
The act-type \lstinline+make-request+ is also defined, with the effect of `creating' an instance of a request, indicating that a specific request has been made by an actor.
The \lstinline+lawful-request+ fact-type determines whether a created request is considered lawful.
The \lstinline+process+ act-type models the actual act of data processing and its condition determines that processing is only permitted when the corresponding request is lawful.
In other words, permission to process must first be explicitly requested, as is common in access and usage control models.
{
\begin{lstlisting}[caption={Specification of the act of processing and of requesting permission.},label={lst:requests},linerange=18-28]
Fact request        Identified by actor * processing-action * purpose * asset
Fact lawful-request Identified by actor * processing-action * purpose * asset
  Conditioned by request() // only considers created requests

Act make-request Actor actor Related to processing-action, purpose, asset
  Holds when actor
  Creates request()

Act process Actor actor Related to processing-action, purpose, asset 
  Holds when actor
  Conditioned by lawful-request()
\end{lstlisting}}\noindent%
The \lstinline+Holds when+ clause of both act-types determines that only actors (controllers or processors) can perform (a request for) processing.
Note that in the expressions \lstinline+request()+ and \lstinline+lawful-request()+ the components of the two referenced relations are omitted.
The missing components have been left implicit and are automatically filled in by the eFLINT interpreter by writing down their names as variables, e.g., \lstinline+request(actor,processing-action,purpose,asset)+.
The implicit variables are in scope as the \lstinline+make-request+ type has the same components, named identically, as \lstinline-request-.
The effect is that executing an instance of \lstinline+make-request+ results in creating an instance of \lstinline-request- with the same components.

The \textit{lawful-request} predicate is defined in the fragment given below, implementing \Cref{eq:specific-of,eq:compatible-with}.
\Cref{eq:base} is not implemented and is redundant (discussed in \Cref{sec:semantics}) as the `specific of' relation is implemented as a reflexive relation (see \Cref{lst:specificity}).
The requirement that a request must exist before it can be considered lawful has been encoded in \Cref{lst:requests}.
{
\begin{lstlisting}[caption={Specification implementing \Cref{eq:specific-of,eq:compatible-with}.},label={lst:lawful-request},linerange=34-46,deletekeywords={with}]
Extend Fact lawful-request
  Holds when prerequisite-of(processing-action, purpose)
          && specific-of(purpose, purpose')
          && legal-basis(controller, purpose', asset)
          && processor-for(actor, controller, purpose')

Extend Fact lawful-request
  Holds when prerequisite-of(processing-action, purpose)
          && sufficiently-specific(purpose)
          && compatible-with(purpose, purpose')
          && legal-basis(controller, purpose', asset)
          && processor-for(actor, controller, purpose')
          && has-been-informed(subject, controller, purpose)
\end{lstlisting}}\noindent%
The variable \lstinline+purpose'+ is not mentioned in the definition of \lstinline+lawful-request+ and is thus an existentially qualified variable in both rules.

The following fragment defines the abstract $\textit{legal-basis}$ relation alongside the six different claims for legal basis (see \Cref{def:legal-basis}) that concretise the abstract relation.
{
\begin{lstlisting}[caption={Specification of the predicates related to legal basis.},label={listing:gdpr-concepts},linerange=49-55]
Fact legal-basis                      Identified by controller * purpose * asset
Fact legal-basis-consent              Identified by controller * purpose
Fact legal-basis-contract             Identified by controller * purpose
Fact legal-basis-legal-obligation     Identified by controller * purpose
Fact legal-basis-vital-interests      Identified by controller * purpose
Fact legal-basis-public-interest      Identified by controller * purpose
Fact legal-basis-legitimate-interest  Identified by controller * purpose
\end{lstlisting}}\noindent
The asset component of \lstinline+legal-basis+ is needed to determine which subjects are involved, as determined by the \lstinline+subject-of+ relation (as shown by \Cref{lst:legal-basis-rules} given below).

The following relations are needed for when there is a claim to consent or the existence of a contract as a legal basis, denoting which subjects have given consent and whether a contract exists between subject and controller.
{
\begin{lstlisting}[caption={Relations needed to derived legal basis in case of consent or contract.},label={listing:gdpr-concepts},linerange=66-68]
Fact has-been-informed Identified by subject * controller * purpose
Fact consent-given     Identified by subject * controller * purpose
Fact contract          Identified by subject * controller * purpose
\end{lstlisting}}\noindent
Both relations have a purpose component as consent is given for a specific purpose and a contract is formed for a specific purpose.
Additional derivation rules (not shown) are associated with the above type-definitions to ensure that consent and contracts that exist for a purpose also exist for all purposes that are more specific.

The fragment below gives an eFLINT definition to the inference rules of \Cref{eq:legal-basis-consent,eq:legal-basis-legitimate-interest} (lines 1 and line 3) alongside the other rules for inferring legal basis omitted in \Cref{sec:semantics}.
{
\begin{lstlisting}[caption={Definition of inference rules and conditions for deriving legal basis.},label={lst:legal-basis-rules},linerange=74-84]
Extend Fact legal-basis Holds when legal-basis-consent()
  && (Forall subject: consent-given() When subject-of())
Extend Fact legal-basis Holds when legal-basis-legitimate-interest()
Extend Fact legal-basis Holds when legal-basis-contract()
  && (Forall subject: contract() When subject-of())
Extend Fact legal-basis Holds when legal-basis-legal-obligation()
Extend Fact legal-basis Holds when legal-basis-vital-interests()
Extend Fact legal-basis Holds when legal-basis-public-interest()
Extend Fact legal-basis 
  Conditioned by sufficiently-specific()
              && (Forall subject: has-been-informed() When subject-of())
\end{lstlisting}}\noindent
The last line in the fragment above specifies that a claim to a legal basis can only be valid if the purpose mentioned in the claim is sufficiently specific.
The \lstinline-Conditioned by- clause introduces a derivation condition, ensuring that an instance of \lstinline+legal-basis+ can only be derived if the condition holds.
The condition is written this way instead of adding it explicitly to each derivation rule, making the code more concise and easier to maintain, as the condition does not need to be specified in each derivation rule independently (as was done in \Cref{eq:legal-basis-consent,eq:legal-basis-legitimate-interest}).
Note that in the expressions of the \lstinline-Holds when- clauses, the components of the mentioned relations are left implicit. 
For example, the expression \lstinline+subject-of()+ is implicitly equal to \lstinline+subject-of(subject, asset)+, thus referring to the asset for which a legal basis is being established.

The predicates and relations concerning specificity of purpose are defined in the fragment below.
The \lstinline+Placeholder+ declarations introduces new variable names as a convenience that can be used in type-definitions to refer to, in this case, instances of the \lstinline+purpose+ type.
{
\begin{lstlisting}[caption={Specification of predicates and relations concerning specificity.},label={lst:specificity},linerange=87-95]
Fact sufficiently-specific Identified by purpose
Placeholder specific-purpose For purpose
Placeholder generic-purpose  For purpose
Fact specific-of Identified by specific-purpose * generic-purpose Holds when
  specific-of(specific-purpose,purpose) && specific-of(purpose,generic-purpose)
 ,specific-purpose == generic-purpose // reflexivity and transitivity (above)
Extend Fact purpose Derived from sufficiently-specific.purpose
                               , specific-of.specific-purpose
                               , specific-of.generic-purpose
\end{lstlisting}}\noindent
In the rule for transitivity, the variable \lstinline-purpose- is an existentially qualified variable as it is not named in the type-definition of \lstinline+specific-of+.
The last three lines ensure that instances of purpose are derived from instances of \lstinline+sufficiently-specific+ and \lstinline+specific-of+ which in turn are given as qualifications.

The fragment below defines the \lstinline+processor-for+ predicate and the \lstinline+dpa+ qualification together with derivation rules implementing \Cref{eq:controller-processor,eq:processor-processor}.
{
\begin{lstlisting}[caption={Specification of \Cref{eq:controller-processor,eq:processor-processor}.},label={lst:processor-for},linerange=98-101]
Fact processor-for Identified by actor * controller * purpose
Fact dpa           Identified by controller * processor * purpose
Extend Fact processor-for Holds when actor == controller
Extend Fact processor-for Holds when dpa()
\end{lstlisting}}\noindent

The implementation of the ontology and its semantics is completed by \Cref{lst:prereq,lst:compatible-with} defining the `prerequisite of' and `compatible with' relations.
{
\begin{lstlisting}[caption={Specification of `prerequisite of'.},label={lst:prereq},linerange=106-107]
Fact prerequisite-of Identified by processing-action * purpose
Extend Fact processing-action Derived from prerequisite-of.processing-action
\end{lstlisting}}\noindent
{
\begin{lstlisting}[caption={Specification of `compatible with'.},label={lst:compatible-with},linerange=110-111]
Placeholder compatible-purpose For purpose
Fact compatible-with Identified by compatible-purpose * purpose
\end{lstlisting}}\noindent
The next subsection demonstrates the implementation with an example application.

\subsection{Example Application}
\label{sec:impl-example}
\begin{figure}
    \centering
    \includegraphics[width=0.85\textwidth]{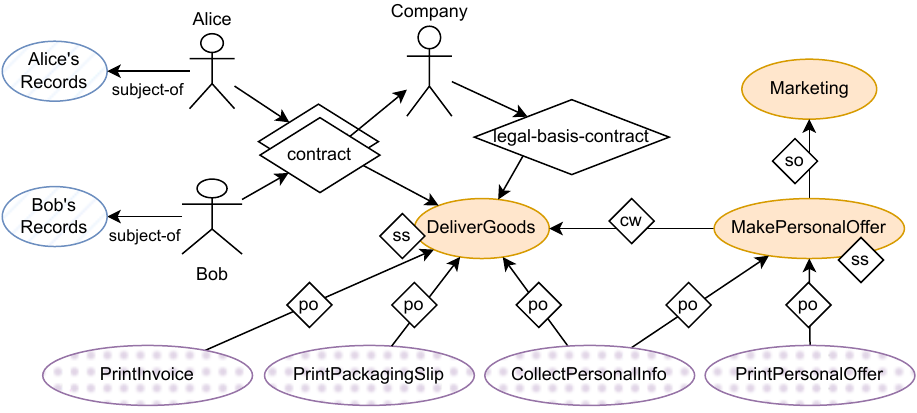} 
    \caption{Example of a purpose-graph instantiating concepts and relations of the ontology in \Cref{fig:ontology}. The names of the relations `prerequisite of' (po), `compatible with' (cw), `specific of' (so) and the predicate `sufficiently specific' (ss) have been abbreviated. Purposes are ellipses with filled background, processing actions are dotted and assets are dashed. The compatibility of MakePersonalOffer with DeliverGoods can be contested.}
    \label{fig:example_graph}
\end{figure}
The example describes a company that delivers goods to customers.
A purpose graph is visualised in \Cref{fig:example_graph}.

This simple example includes two data subjects (Alice and Bob) and one controller (Company) and no third party processors.
The controller collects personal information directly from the data subject for the purpose of delivering goods to their addresses.
The legal basis for the collection can be found in the contract between the data subjects and the controller (\artmm{6}{1}{b}).
After collecting the personal data, the controller performs (further) processing to print the packing slip, an invoice, and a personalised offer.
The example generates at least the following interesting scenarios:
\begin{enumerate}[(a.)]
    \item The processing actions that are prerequisites of delivering goods are lawful, for each individual subject, if a contract exists with that subject and for that purpose. 
    \item The further processing of the data to print and include a personal offer with the packaged goods may be lawful depending on whether the purpose of making a personal offer is considered to be incompatible with the delivery. 
    \item If, instead, the company asks for consent in order to establish an alternative legal basis, the consent needs to state `making a personal offer' as a purpose and not `marketing' as the latter is not deemed to be sufficiently specific.
\end{enumerate}
%
%These further processing is not based on the consent of the data subject or a legal obligation. Therefore the controller need to analyse the a compatibility using the elements stated in \ref{subsec2.1}.
%
The lawfulness of further processing is determined by comparing the processing purpose, stated with the processing action in a request, with the purpose of the legal basis. 
%
% The connection between the purpose for which the data was collected and the printing of packing slips and invoices is clear, i.e., delivering goods in accordance to the contract.
%
The analysis of the lawfulness of making a personalised offer is not straightforward; this purpose is not a more specific purpose than the delivery of the goods (instead it is a specific kind of marketing) and whether the purpose is compatible with the delivery of goods can be debated.
%
% The collected personal information is not of a special nature and implications for the data subject are minimal.
%
% When a customer receives packaged goods as expected, they can simply ignore any personal offer included in the package without much consequence.
%
%Because this case is fictional we assume that appropriate measures are taken such as the proposed model which prevents further processing without a legal basis.
%
In this subsection we will demonstrate automated reasoning with the example purpose graph, showing the effects on the reasoning outcome when small modifications are made to the graph, corresponding to the above scenarios.
%

% The weighing of these elements shows that the further processing of printing the packing slip and invoice is clearly compatible with the original purpose.
%
%For the sake of the example we conclude that the inclusion of a personalised offer is incompatible with the original purpose because the link between the original purpose and the further processing is not strong enough.
%
%In the section an implementation and instantiation of the proposed model will be shown using this fictional case.

The following code fragment adds the assets and associated `subject of' edges as facts to the knowledge base of the reasoner.
\begin{lstlisting}[caption={Statements asserting the truth of instances in the purpose graph related to assets and subjects (corresponding to the Collect capability, see \Cref{tab:policy-capabilities}).}]
+asset(AlicesRecords). +subject-of(Alice, AlicesRecords).
+asset(BobsRecords).   +subject-of(Bob, BobsRecords). 
\end{lstlisting}

The code fragment below adds processing actions and purposes by (only) specifying the relations between processing actions and purposes and the predicate determining which purposes are sufficiently specific.
Note that the `compatible with' edge shown in \Cref{fig:example_graph} is not included in this fragment.
\begin{lstlisting}[caption={Statements asserting the truth of instances in the purpose graph related to processing actions and purposes (corresponding to the Qualify capability, see \Cref{tab:policy-capabilities}).},label={lst:example-qualify}]
+specific-of(MakePersonalisedOffer, Marketing).
+sufficiently-specific(MakePersonalisedOffer).
+sufficiently-specific(DeliverGoods).
+prerequisite-of(CollectPersonalInfo, DeliverGoods).
+prerequisite-of(CollectPersonalInfo, MakePersonalisedOffer).
+prerequisite-of(PrintPackingSlip, DeliverGoods).
+prerequisite-of(PrintInvoice, DeliverGoods).
+prerequisite-of(PrintPersonalisedOffer, MakePersonalisedOffer).
\end{lstlisting}%
Recall that the instances corresponding to the processing actions and purposes are derived automatically (see previous subsection) and therefore do not need to be given.

\paragraph{Scenario (a)}
The fragment below shows the claim for a legal basis based on contracts with the subjects, as well as the qualification that such contracts exist.
\begin{lstlisting}[caption={Statements asserting the truth of instances in the purpose graph related to legal basis (corresponding to the Control capability, see \Cref{tab:policy-capabilities}).},label={lst:request1}]
+legal-basis-contract(Controller, DeliverGoods).
+contract(Bob, Company, DeliverGoods). +contract(Alice, Company, DeliverGoods).   
\end{lstlisting}%

The provided information has fully populated the purpose graph as shown in \Cref{fig:example_graph}.
The following statements show how a request is made, subsequently tested for lawfulness, and processed\footnote{Note that the actual processing occurs external to the reasoner. The final statement of \Cref{lst:request1} instead reflects the \textit{observation} or \textit{qualification} that processing has occurred.}.
The company is the actor making the request and performing the processing, as no third party processor are present in this example.
\begin{lstlisting}[caption={A code fragment querying whether the printing of an invoice based on Bob's record is lawful for the purpose of delivering goods.}]
make-request(Company,PrintInvoice,DeliverGoods,BobsRecords).   //no violation
?lawful-request(Company,PrintInvoice,DeliverGoods,BobsRecords).//query succeeds
process(Company,PrintInvoice,DeliverGoods,BobsRecords).        //no violation
\end{lstlisting}
The output of a statement is summarised in the comment directly following the statements.
Removing the qualification that a contract for delivering goods exists between the Company and Bob makes the query fail and the final statement raise a violation.
Note that in a more realistic example we can specify the processing purpose `deliver goods' in more detail and relate it to, for example, an purchasing order.
In this case, the purpose can be related to Bob, through the order, making it possible to establish a connection between the asset and the processing purpose of a request.
This way it is possible to enforce that delivering goods to Bob does not allow processing the data of Alice (despite the existence of a contract for delivering goods to Alice as well).

\paragraph{Scenario (b)}
The following fragment shows that printing a personal offer is not considered lawful as the processing purpose `make personal offer' is deemed to be incompatible with delivering goods (as determined by omitting the `compatible with' edge between the two purposes).
\begin{lstlisting}[caption={A code fragment querying whether the printing of a personal offer based on Bob's record is lawful for the purpose of making a personal offer. The subsequent statement raises a violation as the processing is not considerd lawful.},label={lst:scenariob}]
make-request(Company,PrintOffer,MakePersonalOffer,BobsRecords).   //no violation
?lawful-request(Company,PrintOffer,MakePersonalOffer,BobsRecords).//query fails
process(Company,PrintOffer,MakePersonalOffer,BobsRecords).        //violation
\end{lstlisting}
Had the statement \lstinline[deletekeywords={with}]~+compatible-with(MakePersonalOffer,DeliverGoods)~ been included in \Cref{lst:example-qualify}, then the query would have succeeded.

\paragraph{Scenario (c)}
Instead of relying on a negative incompatibility assessment (resulting in a \lstinline[deletekeywords={with}]+compatible-with+ qualification), the company can also ask for consent to make personalized offers, as many companies do in their webshops. 
The company can claim an additional legal basis by executing the assertion \lstinline+legal-basis-consent(Company,Marketing)+. 
However, the query of \Cref{lst:scenariob} would still fail because (1) the marketing purpose is not sufficiently specific and (2) consent has not yet been given by Bob.

The statement \lstinline~+consent-given(Bob,Company,Marketing)~ asserts that Bob has given consent for the purpose of marketing.
Note that the assertion that Bob gives consent for making personalised offers will be derived from this assertion, as making a personalised offer is more specific than marketing.
However, as discussed in \Cref{sec:purpose-limitation}, consent should be asked for a purpose that is sufficiently specific.
If marketing was considered to be sufficiently specific, by adding the assertion \lstinline~+sufficiently-specific(Marketing)~, then the query in \Cref{lst:scenariob} would succeed.

Instead, the company claims an additional legal basis with the following fragment. 
\begin{lstlisting}[caption={The assertion of consent for making personalised offers as an additional legal basis.}]
+legal-basis-consent(Company,MakePersonalOffer).
\end{lstlisting}
If Bob gives consent, as specified by the fragment below, then the processing action of printing a personal offer is considered lawful and the query of \Cref{lst:scenariob} succeeds.
\begin{lstlisting}[caption={Statement asserting consent has been given (the Consent capability, see \Cref{tab:policy-capabilities}).}]
+consent-given(Bob,Company,MakePersonalOffer).
\end{lstlisting}

The discussion of the scenarios demonstrates the importance of qualifications.
The next section reflects on the soundness of qualifications, observing that ``garbage in is garbage out'', but also that through making qualifications explicit, the decision making process has become more transparent and accountable.
The next section also discusses who is responsible for providing which qualifications, i.e., which parts of the purpose graph are provided by which stakeholders and software components of a data exchange system.
As such, the next section describes how the implementation of this section can be integrated into (existing) data exchange systems.

\section{Integration}
\label{sec:architecture}
\label{sec:integration}
The concepts and relations of the ontology in \Cref{fig:ontology} form the basis of automated, normative reasoning as formalised in \Cref{sec:semantics} and implemented and demonstrated in \Cref{sec:implementation}.
To derive conclusions about the lawfulness of specific processing actions, concrete {instances} of concepts and relations of the ontology need to be given.
For example, \Cref{def:request} and \Cref{eq:base} show that for a specific processing request, knowledge is required about the involved subjects and the specific legal basis being claimed.
The required instances of these and the other relations are collected in a so-called purpose graph, an extended version of the purpose graphs proposed by~\cite{byun2005purpose}.
An example purpose graph has been given in \Cref{fig:example_graph}.

The first goal of this section is to describe in general terms how a purpose graph is constructed and applied, given the challenge that responsibilities and informational content is distributed among actors within a (distributed) data processing infrastructure (\Cref{sec:roles}). 
The second goal is to describe and distribute technical roles charged with enforcing lawfulness of processing among components of a distributed system (\Cref{sec:technical-architecture}).
The result is a collection of high-level architectural patterns that can be applied in the development of a distributed data processing system. 
The architectural patterns are described in terms policy administration roles and capabilities, technical policy enforcement roles, delegation archetypes, and flows of information.
Finally, administration and enforcement roles come together in a use case in \Cref{sec:integration-example}.

\subsection{Policy Administration Roles and Capabilities}
\label{sec:roles} 
\begin{figure}
    \centering
    \includegraphics[width=0.35\linewidth]{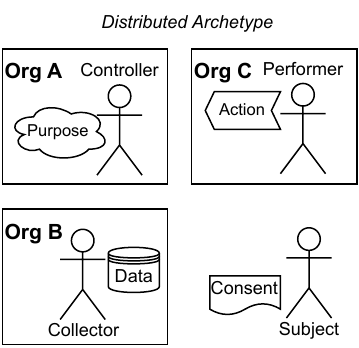}
    \hspace{4em}
    \includegraphics[width=0.35\linewidth]{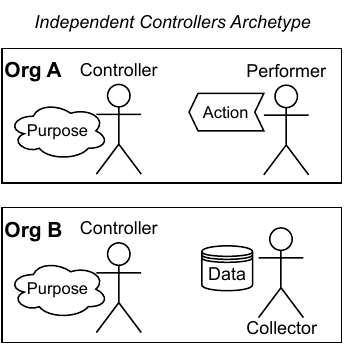}
    \caption{For a single processing action, this diagram identifies the party (organisation, actor) performing the action (C, Performer), the party collecting the data required for that action (B, Collector), and the party determining the purpose of the processing (A, Controller). If consent is required for processing, then this is provided by the Subject. The actor-labels correspond to the policy administration roles defined in \Cref{tab:policy-capabilities}. As depicted on the left, all responsibilities related to the processing action are distributed across different organisations. In other processing archetypes, one organisation plays multiple roles, as defined in \Cref{tab:policy-roles}. The Independent Controllers archetype is on the right. Diagrams of other archetypes are formed by fusing the boxes of the Distributed processing archetype.}
    \label{fig:distributed-archetype}
\end{figure}
% %
% \begin{figure}
%     \centering
%     \includegraphics[width=0.75\linewidth]{images/policy_roles.pdf}
%     \caption{Roles and their capabilities to submit policy fragments or reasoning requests to a normative reasoning component.}
%     \label{fig:policy-roles}
% \end{figure}
% %
\begin{table}[tb]
    \centering
    \begin{tabular}{|l|l|l|}
    \hline
      \textbf{Capability} & \textbf{Purpose-graph contributions} & \textbf{Assigned to}\\
      \hline
    Control & legal-basis, dpa, has-been-informed, contract(s) (if applicable) & Controller,\\
    & & Authority\\\hline
    Qualify & prerequisite-of, compatible-with, specific-of, sufficiently-specific & Controller,\\
    & & Authority\\\hline
    Collect & asset(s), subject-of & Collector\\\hline
    Perform & request & Performer\\
    & & Collector\\\hline
    Consent & consent-given (including withdrawal of consent) & Subject\\
    \hline
    \end{tabular}
    \caption{Definition of policy administration capabilities providing policy information in the form of purpose graph components. The capabilities are assigned to the policy administration roles Controller, Collector, Performer, Authority, and Subject. Perform is assigned to both Performer and Collector as performing an action and collecting data are both considered processing.}
    \label{tab:policy-capabilities}
\end{table}
\begin{table}[tb]
    \centering
    \begin{tabular}{|l|r|l|}
    \hline
       \textbf{Processing Archetype} & \textbf{Organisation} & \textbf{Policy Administration Roles}\\
      \hline
     No Delegation & Controller & Controller, Collector, Performer\\\hline
     Delegated Action & Controller & Controller, Collector\\
     & Performer & Performer\\\hline
     Delegated Processing & Controller & Controller\\
      & Performer & Collector, Performer\\\hline
     Delegated Collection & Controller & Controller, Performer\\
      & Collector & Collector\\\hline
     Distributed & Controller & Controller\\
      & Collector & Collector \\
      & Performer & Performer \\\hline
    Independent Controllers & Controller A & Controller, Collector \\
    & Controller B & Controller, Performer \\\hline
    \end{tabular}
    \caption{An assignment of policy administration roles to organisations in several archetypical scenarios related to a single processing action. In every scenario, there is a controller organisation with at least the Controller role. An organisation with only one role in an archetype is named after that role in the archetype.}
    \label{tab:policy-roles}
\end{table}

In this and the following subsection we will analyse unitary `scenarios' in which a single processing action is applied to a collected dataset for the purposes of a controller. 
The responsibilities and contributions in such scenarios of different organisations and their actors are summarised and visualised in \Cref{fig:distributed-archetype}.
In \Cref{sec:integration-example}, these types of scenarios are considered as the building blocks of larger `use cases'.

Which policy information -- collection of components of the purpose graph -- is provided by which organisation is dependent on the details of a scenario. 
We define several \textit{policy administration roles} that each have different capabilities regarding the policy information they contribute. 
\Cref{tab:policy-capabilities} summarises which policy contributions are made by which of the involved roles by assigning elements of the ontology to `capabilities' and by assigning capabilities to the roles.
The roles are subsequently assigned to organisations in \Cref{tab:policy-roles} according to `delegation archetypes', inspired by the data exchange archetypes of~\cite{shakeri2020} and~\cite{zhang2019}.

As examples, consider the following scenarios.
In the first scenario, a controller collects the data and delegates (only) the application of an analysis algorithm to a processor (the Delegated Action archetype, \Cref{tab:policy-roles}).
In the second scenario, the controller also delegates the collection of data to the processor (Delegated Processing).
In the first scenario, the controller creates assets by labelling datasets and determines which subjects are identifiable in the labelled datasets (the Collect capability).
In the second scenario, this task is performed by the processor. 
In both scenarios, the processor performs a processing action, the analysis algorithm, for which lawfulness needs to be established by making a request (the Perform capability).
Since data collection is processing according the GDPR, both described scenarios involve two processing actions for which lawfulness needs to be established.
That is, both reading and writing a data resource needs to be subjected to access control and trigger a request for a decision in its lawfulness. 

In a third scenario, two controllers collaborate with one controller performing the Collector role and the other the Performer role (Independent Controller archetype).
The two organisations are both considered controllers if they both, independently, determine the legal basis and purpose for their contribution to the collaboration.
An example of such a use case is given in \Cref{sec:integration-example}.

The primary task for any controller is to establish the legal basis for the processing.
Thus, in any scenario, a controller has the Controller role assigned.
If the legal basis is consent, authorisation requires that consent was given by all\footnote{Note that our ontology also supports representing actors at a coarser granularity than individual persons, which may be more practical in some contexts.} subjects (the Consent capability).
If the legal basis is the fulfilment of a contract, lawfulness requires that indeed such a contract exist between the controller and data subject(s).

Part of the role of the controller is also to determine the `prerequisite-of' relation between actions and purposes, and the compatibility and specific-of relation between purposes (the Qualify capability).
Another element of establishing specificity (and the Qualify capability) is determining whether the purpose for processing is sufficiently specific for the data subject(s) to understand the processing that will take place.
As shown in \Cref{tab:policy-capabilities}, both the Control and Qualify capabilities are also associated with the Authority role.
Ex-post scenarios with authorities are discussed in \Cref{sec:ex-post-scenarios}.

Each of the scenarios in \Cref{tab:policy-roles} involves exactly one processing action and data collection (also considered processing).
And, in the case this action is delegated to a processor, involves a bilateral processing agreement between controller and processor.
In many realistic use cases, however, an agreement between multiple parties is formed.
Multi-party agreements can involve one or more controllers and one or more processors. 
Multiple organisations can also decide to form a `joint controllership'. 
In the first case, the Performer role needs to be assigned multiple times to multiple organisations, once for every processing action, such that each of the organisations can make processing requests.
The Collector role is assigned to each organisation collecting (personal) data in multi-party scenarios.
%
% Section~\ref{sec:infrastructure} discusses several data exchange topologies and possible assignments of roles.
%
The investigation of joint controllership is left as future work.

\paragraph{Reflections on the need for qualifications}
Qualifications are needed by the normative reasoning process described in \Cref{sec:semantics,sec:implementation} to determine the lawfulness of a processing action.
An access/usage control implementation that bases its authorisations on the proofs constructed through the normative reasoning process thus forces the controller to make certain qualifications explicitly.
Where desired, (digital) signatures on elements of the purpose graph can be used to confirm the identity of the party providing the elements. 
%
%Further more, manual signatures can be collected to ensure that a person has made the determination, rather than a software component. 
%
For example, a legal analyst of the controller signs the claimed legal basis and a data subject signs for giving consent.
% %
% However, a ruling on the level of specificity is ultimately made by the privacy authority.
% %
% More detail on how this can work in practice is provided in \Cref{sec:usage-scenarios} in which usage scenarios are described in more detail.
%

%
In general, the normative reasoning of our proposal is as sound as the input provided to the reasoner.
A malicious user can game the system by simply omitting instances of the `subject-of' relation, by setting all purposes as compatible, or stating that all actions are prerequisites, etc.
Therefore, audits by privacy authorities are still required, even if our model is adopted by fully secure implementations.
And as we shall see in \Cref{tab:ex-post-policy-roles}, the Qualify capability is assigned to both the Controller and the Authority role in auditing scenarios, making explicit the details of any conflicting qualifications between controller and privacy authority.
Similarly, data subjects can make a request for information to determine the legal bases claimed by the controller for processing actions on their personal data.
Based on their own qualifications, e.g., that consent was withdrawn, data subjects can make there own assessments.
These ex-post scenarios are described further in \Cref{sec:usage-scenarios}.

The strength of our proposal is that qualifications are made explicit as part of normative reasoning, and can therefore also be logged and made available to promote transparency, accountability and auditability. 
Moreover, our approach makes it possible for access/usage control decisions to be based on legal qualifications rather than on system-level policies.
The legal qualifications can be set by a privacy expert, whereas system-level policies are typically set by a system administrator.
These organisational roles are commonly associated with strongly divergent responsibilities and competences.
And even in organisations where legal experts and system administrators work closely together, their collaboration may be prone to error and introduces overhead.
A central hypothesis to this work is that, by reducing the conceptual gap between the legal/privacy domain and the computational domain, the risk of non-compliance and the costs of maintenance are reduced.

\subsection{Technical Enforcement Roles}
\label{sec:technical-architecture}
\begin{figure}
    \centering
    \includegraphics[width=0.6\linewidth]{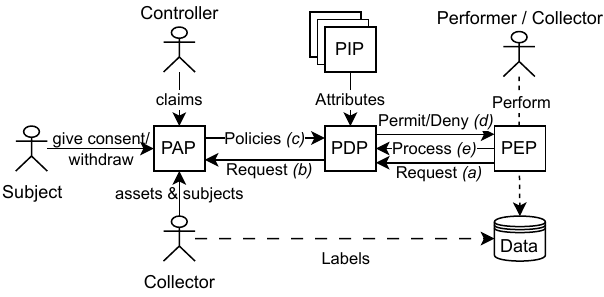}
    \caption{An extension of (a simplified version) of the architectural pattern in the XACML standard that introduces the purpose graph and the procedure for determining whether a processing request is lawful. The actors interacting with the architecture are labelled by their capabilities (see \Cref{tab:policy-capabilities}). The italicised letters show the order of messages. Additional policies can be sent alongside the purpose graph. The `Process' message is sent only when authorisation has been given and processing occurs.}
    \label{fig:xacml-architecture-administration}
\end{figure}
The previous subsection discussed the mapping of policy administration roles to organisations within delegation archetypes for processing scenarios. 
This subsection discusses the mapping of technical, \textit{policy enforcement roles} to organisations.

The technical roles are taken from the XACML architecture as described in \Cref{sec:background-xacml}.
The XACML architecture is chosen as a de facto standard for access and usage control systems.
An added benefit of our choice is that our approach can be implemented as an extension of existing systems implementing the XACML architecture, of which many exist in practice.
\Cref{fig:xacml-architecture-administration} extends the XACML architecture of \Cref{fig:xacml-roles}, associating policy administration roles (as actors) with some of the technical roles PEP, PDP, PAP, and PIP.
The interaction protocol between the components of the architecture is extended with the additional communication (edge \textit{e}) between PEP and PDP, indicating that a processing action has been performed.
The policies sent by the PAP to the PDP (edge \textit{c}) include at least the case-generic specification (the code fragments given in \Cref{sec:implementation-semantics}) and the purpose graph (of which example code fragments were given in \Cref{sec:impl-example}) on top of any additional access and usage policies.
The inclusion of the case-generic specification increases the adaptability of the system as any updates to the ontology and semantics can be applied dynamically as the set of applicable policies is determined on a per-request basis.

In \Cref{fig:xacml-architecture-administration}, most policy administration roles are connected to the policy administration point (PAP).
The request stems from the policy enforcement point, however. 
Through the policy decision point (PDP), the request is communicated to the PAP to determine which purpose graph held by the PAP is relevant to the request, if multiple are maintained.
The purpose graph is communicated to the PDP which then determines the validity of the request through normative reasoning, as formalised and operationalised in \Cref{sec:semantics,sec:implementation} respectively.
The diagram suggests that the actor with the Collector role is the actor that has physically collected the data (perhaps on behalf of another actor, the controller).
As collecting personal data is also considered processing personal data according to the GDPR, authorisation for this action should also be requested.
The attributes provided by the policy information point (PIP) can be used by the PDP in conjunction with the policies given by the PAP in addition to the purpose graph (a topic further discussed in \Cref{sec:discussion}). 
This way, purpose-based normative reasoning is an orthogonal extension to existing access and usage control models.

The GDPR requires that lawfulness is reviewed periodically. 
This requirement is partly satisfied by our proposal as the lawfulness of individual processing requests are reviewed when they occur. 
A review of the (qualifications in the) purpose graph can be enforced by implementing the PAP such that it discards (elements of) the purpose periodically after a configurable amount of time.

\begin{figure}
    \centering
    \includegraphics[width=0.7\linewidth]{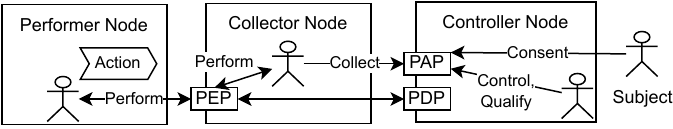}
    \\[1.5em]
    \includegraphics[width=0.65\linewidth]{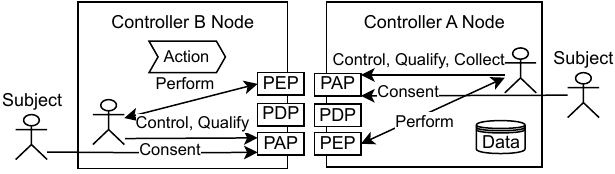}
    \caption{A self-governed peer-to-peer data exchange system adhering to the Distributed processing archetype (top) and Independent Controllers archetype (bottom). The data exchange is self-governed in the sense that the policy enforcement roles are implemented by the organisations themselves.}
    \label{fig:self-governed}
\end{figure}
\begin{figure}
    \centering
    \includegraphics[width=0.8\linewidth]{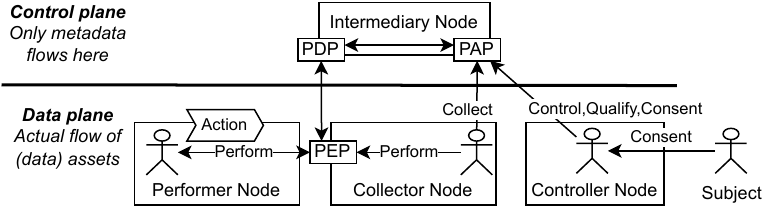}
    \\[1.5em]
    \includegraphics[width=0.65\linewidth]{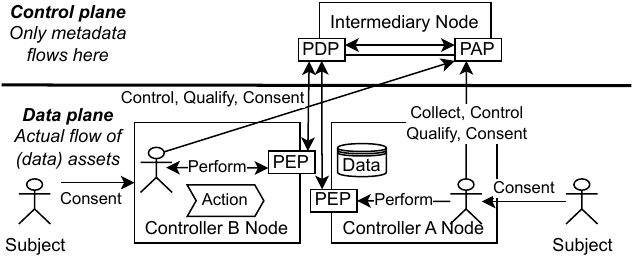}
    \caption{An intermediary-governed peer-to-peer data exchange system adhering to the Distributed processing archetype (top) and Independent Controllers archetype (bottom). The intermediary implements the PDP and PAP technical roles, relieving the controller(s) of this task. As in \Cref{fig:self-governed}, the Consent capability is for the subject to inform the controller. The controller collects consent and informs the Intermediary. The distinction between control plane and data plane is made in reference to the AMdEX reference architecture (see~\cite{amdex-ra}) in which the role of intermediary is discussed more extensively.}
    \label{fig:intermediary-governed}
\end{figure}
The diagrams of \Cref{fig:self-governed,fig:intermediary-governed} show different ways in which the technical roles of XACML and the policy administration roles can be mapped unto a data exchange system.
In these diagrams, data exchange systems contain `nodes' that correspond to the software infrastructure of different organisations according to the Distributed processing archetype.
The Distributed processing archetype is used as in this single controller archetype, the most responsibilities are delegated by the controller.
Similar diagrams for the other single controller archetypes can be obtained by simply merging nodes, e.g., merging the Performer and Collector node to obtain the Delegated Processing archetype or merging the Collector Node and the Controller Node to obtain the Delegated Action archetype.
There are two controller nodes in the diagram for the Independent Controller archetype.
Controller $A$ node implements the Collector role and a PEP for ensuring the lawfulness of collecting data. 
Controller $B$ node implements the Performer role and a PEP for ensuring the lawfulness of the performed processing action.
Both implement the Controller role for their respective processing and, if necessary, the Consent capability.
Unless delegated to an intermediary, independent controllers both implement a PAP and PDP.
In scenarios within the Independent Controller archetype, no intercommunication between organisations is necessary.

The difference between \Cref{fig:self-governed} and \Cref{fig:intermediary-governed} is that in the latter, an intermediary is introduced as an additional organisation responsible for implementing the PDP and PAP.
In this variant, the Collect, Control, and Qualify capabilities are to inform the Intermediary rather than the Controller.
Similarly, the intermediary can also be informed about consent.
Note that in contexts where subjects' consent and contribution to a dataset is considered sensitive information, the subject-of relation (provided by the Collector) and the consent claim (provided by Controller) can be provided in aggregate form. 
For example, a dataset can be labelled as identifying ``female patients over 50'' and the consent provided by the controller can state that all patients in this cohort have given consent. 
In these cases, the Controller is still responsible of collecting and retaining the consent by the individuals.

Analysing the diagrams for different archetypes, we observe the following:
\begin{itemize} 
    \item An organisation holding data implements Collect, Perform, and a PEP.
    \item An organisation executing a processing action implements Perform.
    \item An independent controller with the Performer role implements a PEP.
    \item A Performer organisation connects Perform to a remote PEP.
    \item A Collector organisation connects Perform to the local PEP.
    \item A controller connects Perform to the local PEP.
    \item A controller implements (at least) the Control and Qualify capabilities.
    \item A controller facilitates the Consent capability (i.e., collects consent).
    \item A controller implements a PDP and a PAP, unless delegated to an intermediary.
    \item A processor has their actions authorised by an external decision-making process.
\end{itemize}
%
%The penultimate invariant is a necessary constraint as consent can be considered sensitive information.
%
%Moreover, the Controller is responsible for proving consent has been given where needed (as described in \Cref{sec:purpose-limitation}). 
%
%The last invariant is not a constraint, i.e., a data intermediary may consist of multiple components and multiple organisations.
%
%An analysis of this case is left out of scope.
%

The assignment of technical roles to system components is prescriptive in a minimal sense and is not intended to describe complete system implementations.
Firstly, \Cref{fig:self-governed,fig:intermediary-governed} (and later also \Cref{fig:kpn-actions}) do not show PIPs.
The need for policy information is dependent on the specifics of any additional policies that need to be applied in a specific case (edge \textit{c}, in \Cref{fig:xacml-architecture-administration}).
Instead, the diagrams are intended to describe general patterns, covering many possible cases.
Secondly, additional PAPs, PDPs, and PEPs can be implemented, if so desired.
For example, in an intermediary governed data exchange system connecting independent controllers, all organisations can decide to implement PDPs and PAPs such that decisions and policies are being federated between these components.
This way, more control is retained by the controllers, while the protocol of the bottom diagram in \Cref{fig:intermediary-governed} can still be realised.

\subsection{Role Assignment in Data Processing Use Cases}
\label{sec:integration-example}
%
% \begin{figure}
%     \centering
%     \includegraphics[width=0.6\linewidth]{images/case.pdf}
%     \caption{An example case in which three companies collect (personal) employee data for the purpose of fulfilling an employment contract (1, No Delegation archetype) and share the data with an industry association to fulfil a legal obligation (2, Delegated Processing archetype). The oversight by a privacy of the processing of the industry association (3, Auditing) is also included in this example. The integration  demonstrates an assignment of technical roles (PEP, PDP, PAP) to the organisations involved, as well as the roles and policy capabilities according to the usage scenarios of \Cref{tab:policy-roles}.}
%     \label{fig:mapping-example}
% \end{figure}
%
%
\begin{figure}
    \centering
    \includegraphics[width=0.48\linewidth]{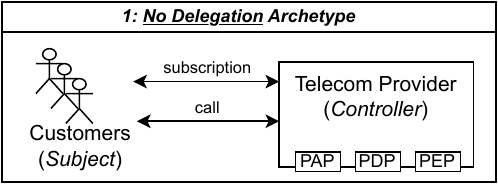}\hfill{}
    \includegraphics[width=0.48\linewidth]{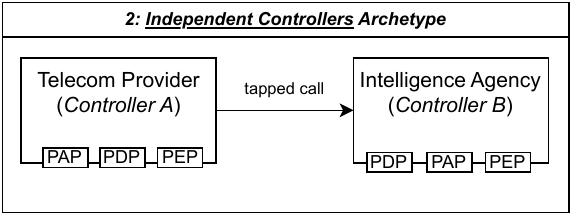}
    \\[1.5em]
    \includegraphics[width=0.75\linewidth]{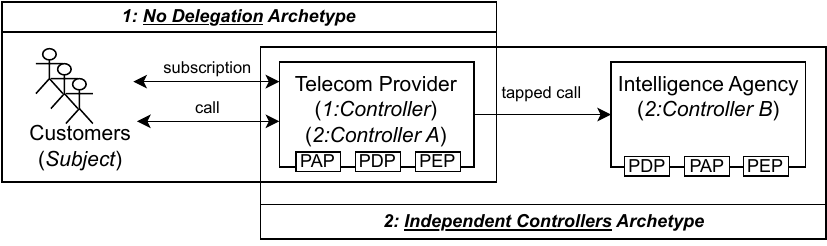}
    \caption{The three diagrams display the assignment of the policy administration roles Controller and Collector and the policy enforcement roles PAP, PDP, PEP in relation to the two processing actions `handling a call' (top left) and `transmitting a tapped call' (top right). In the second scenario, the Telecom Provider hosts the PEP, performing the Collector policy administration role within the Delegated Collection archetype. The PEP and PDP communicate about the lawfulness of a specific tap. The bottom image displays the overarching use case consisting of the two processing actions.}
    \label{fig:kpn-actions}
\end{figure}

The previous two subsections have defined policy administration roles and technical enforcement roles (respectively). 
In both cases, the assignment of roles to organisations and their software components has been described at the level of an individual processing action (together with the collection of the data used by that processing action). 
This subsection describes the assignment of both types of roles in use cases consisting out of one or more such processing actions.
The discussion is centred around the following, practical use case\footnote{The case is real, but the system described in this subsection is not implemented. That is, KPN does not implement the technical roles nor applies normative reasoning as described in this paper.} of the Dutch telecommunications provider KPN and its legal obligation to support wire-taps as part of criminal investigations conducted by authorised law enforcement agencies. 
Besides the GDPR, national and international (EU) directives regarding privacy and electronical communication also apply.

As a telecommunications provider, KPN collects personal data on their customers in order to fulfil a service contract.
This includes personal data necessary for invoicing and meta-data regarding these communications (e.g. recipient's phone number and duration). 
In the first scenario, depicted as an instance of the No Delegation archetype in \Cref{fig:kpn-actions}, we consider the handling of calls.
KPN determines \artmm{6}{1}{b} (`performance of a contract') as the legal basis, with the purpose of `providing a service', for processing the calls.
Applying the mappings of roles discussed in \Cref{sec:roles,sec:technical-architecture}, KPN performs both the Controller, Collector and Performer administration roles (as the Controller of the No Delegation archetype) and would need to implement the PAP, PDP, and PEP technical roles to apply normative reasoning.

In the second scenario, depicted as an instance of the Independent Controllers archetype in \Cref{fig:kpn-actions}, KPN transmits the contents of a phone call as a wiretap to the law enforcement agency. 
As controllers, both organisations determine the legal basis and purpose for processing independently.
The law enforcement agency claims \artmm{6}{1}{e} (`vested authority') with the purpose of conducting a criminal investigation.
KPN claims \artmm{6}{1}{c} (`legal obligation') with the purpose of satisfying a specific legal obligation placed on KPN by national laws.

KPN is the independent controller implementing the Collector administrative role (Controller A of the Independent Controllers archetype) and the ensuing Collect capability, Perform capability and a PEP.
The PEP ensures that specific wiretaps leaving the premises of KPN are lawful by interacting with the local PDP and PAP, the latter of which contains the purpose graph and additional policies established by KPN.

The law enforcement agency is the independent controller implementing the Perform administrative role (Controller B) and the ensuing Perform capability and a PEP. 
The PEP ensures that the wiretaps, collected by the law enforcement agency are lawful.
As is the case for KPN, lawfulness is established by interacting with the local PDP and PAP.

The bottom diagram of \Cref{fig:kpn-actions} gives an overview of the entire use case by overlapping the images of the individual scenarios. 
There is no communication between the software components of both organisations, other than the technical means with which the wiretap is transmitted.
This is consistent with the determination that both organisations are independent controllers and, therefore, are individually responsible for determining the lawfulness of their processing actions.
A processor, on the other hand, would connect to a remote PDP and PAP combination implemented by a controller, or of an intermediary implementing them on behalf of a controller. 

%STEPS
% Introduce the three scenarios (processing actions) one-by-one, each with a picture.
% Mention that self-governed peer-to-peer applies
% Show composite picture with GDPR-role assignment
% Discuss 
%\Cref{fig:mapping-example} gives an example of mapping the technical and administrative roles in a hypothetical case.

\section{Ex-post Normative Reasoning Scenarios}
\label{sec:usage-scenarios}
\label{sec:ex-post-scenarios}
\begin{figure}
    \centering
    \includegraphics[width=.8\linewidth]{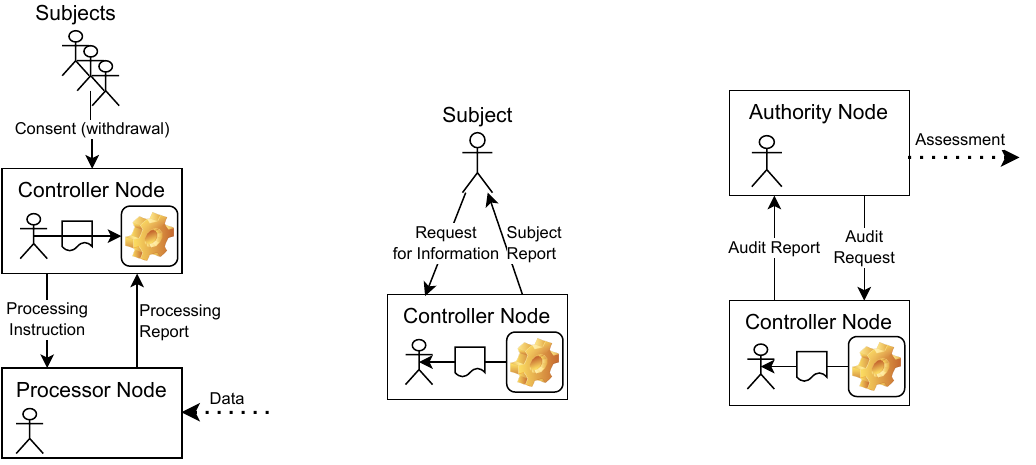}
    \caption{Diagram showing three usage scenario types for normative reasoning: generating processing authorisations (Processing, \Cref{tab:policy-roles}), requests for information by data subjects (Information Request), and inspection by a privacy authority (Auditing). A `node' represents the software infrastructure localised to a particular organisation, possibly offering interfaces to internal actors, external actors and other nodes. In the first scenario, the origin of the data is deliberately undetermined; it may be the controller (No Delegation archetype, \Cref{tab:policy-roles}), the processor itself (Delegated Processing), or a third party processor (Distributed). The gear icon is used to depict the interaction of an internal actor with the normative reasoner in support of fulfilling the represented scenario. In Processing scenarios, information is provided to the reasoner to establish the lawfulness of a particular processing action (as described in \Cref{sec:integration}). In the Information Request and Auditing scenarios information is extracted from the reasoner in order to provide reports to the data subject and authority respectively.}
    \label{fig:usage-scenarios}
\end{figure}
\begin{table}[tb]
    \centering
    \begin{tabular}{|l|r|l|}
    \hline
       \textbf{Scenario type} & \textbf{Role} & \textbf{Capabilities}\\
      \hline
    Information Request & Controller & All\\
    & Subject & Qualify, Consent (possible dispute)\\\hline
    Auditing & Controller & All\\
    & Authority & Qualify, Control (possible dispute)\\
    \hline
    \end{tabular}
    \caption{An assignment of capabilities to roles in ex-post scenarios. The recurrence of certain capabilities across roles shows potential conflicts.}
    \label{tab:ex-post-policy-roles}
\end{table}
\Cref{sec:integration} has primarily focused on the integration of normative reasoning in systems in which lawfulness of a processing action is analysed when the action is about to happen (access control) or is currently happening (usage control). 
However, our model for normative reasoning can also be applied in \textit{ex-post} scenarios.
In this section we discuss two types of ex-post scenarios: (1) a data subject sending a request of information about the usage of their personal data to a controller and (2) a privacy authority requesting an audit from a controller.
Together with the processing scenario type (see \Cref{tab:policy-roles}), these two types of ex-post scenarios are depicted visually in \Cref{fig:usage-scenarios}.
\Cref{tab:ex-post-policy-roles} defines the policy administration roles for these two types of scenarios.

%The primary usage scenario for the conceptual model is to determine whether a processing action can be authorised \textit{prior} to its execution (left-hand side, \Cref{fig:usage-scenarios}), as is typical in a system applying access control.
%
%The model can be used similarly in a system applying usage control, in which case, the determination of lawful processing can also be made continuously, \emph{during} the processing. 
%

%\subsection{Auditing}
%
In the case of ex-post auditing by an internal privacy officer or an external privacy authority, the auditor is interested in, among other things, the processing actions performed by the controller and its processors.
The auditor can dispute the explicitly claimed legal basis for processing as well as other qualifications made by the controller such as whether a purpose is considered sufficiently specific.
%
% (the Control and Qualify capabilities).
%
% If the auditor were to perform normative reasoning as well, the privacy authority can offer its own qualifications to the reasoner, establishing the authority's own view of the processing activities under scrutiny.
%
The construction of an audit report can be partially automated if: (a) the implementation of the data exchange system ensures that processing actions are only performed when lawfulness can be established, (b) the detailed manner in which the lawfulness of a processing action is established is recorded in a persistent manner, and (c) all processing activities are recorded when they occur.
\Cref{sec:integration} has discussed several ways in which (a) can be achieved with our model for purpose-based normative reasoning.
And because our model is based on logical inference, the manner in which lawfulness has been established (requirement (b)) can easily be recorded (as derivation trees).
Requirement (c) is represented by the edge labelled `Process' in \Cref{fig:xacml-roles}. 
A mechanism like the obligation-mechanism found in XACML can be used to implement the requirement, forcing the PEP to inform the PDP\footnote{It may be preferred to introduce a separate component dedicated to persistent logging of reasoning decisions and processing activities, increasing separation of concerns and the robustness of the system.} of processing actions when they are performed.

The information in the derivation tree can be used to produce information for audit reports.
The eFLINT interpreter used for our implementation does not yet record derivation trees, i.e., does not show exactly how facts are derived.
The interpreter does record, and is capable of exporting, all rules, triggers, facts and outputs it encounters
A privacy officer or data protection authority can use the information in the export to audit the data processing that has taken place.
The detailed requirements, mechanics and evaluation of the auditing process are left as future work.
%

%\subsection{Information Requests}
%
The same information produced for auditing can be used as part of the creation of a report requested by a data subject as part of a request for information, a right given to the data subject by \artm{15}{1} of the GDPR.
In particular, the information recorded by the reasoner can be used to determine which assets involving the subject have been processed, for which purpose(s), and which legal basis was claimed.
The information provided to the data subject should be presented in a concise and understandable manner, an additional requirement specified in \art{12} of the GDPR.
The exported information should not be sent to the data subject directly as the format will, in most cases, not be deemed understandable and concise.
The controller should therefore translate the technical information to fulfill their duty under \art{12}.
The detailed requirements, mechanics and evaluation of this process are left as future work.

% The definition of \lstinline-process- given in \Cref{sec:implementation} is extended as follows to log certain information when processing activity is observed.
% %
% % Note that an issue with the code below is that the legal-basis, sufficiently-specific, subject-of relations (etc.) can be modified afterwards, making that insufficient information is recorded to use the reasoner to produce the correct reports.
% % Eg, was consent given at the time the processing happened?
% %
% \begin{lstlisting}[caption={The act-type \lstinline-process- is extended such that it records information about its triggered instances. A var-type is a fact-type for which at most one instance can hold at a time.}]
% Extend Act process
%   Creates processed() // records info about the triggered instance of process
%         , processed-id(processed-id + 1) // id = id + 1
% Fact processed    Identified by processed-id * processing-action * asset * purpose
% Var  processed-id Identified by Int. // at most one instance can hold
% +processed-id(1). // id = 1
% \end{lstlisting}
% %
% In the context of the example of \Cref{sec:impl-example}, the following {instance query} can be used to generate all the processing logs of processing actions touching data for which Bob is the subject.
\section{Discussion}\label{sec:discussion}
This section critically reflects on the approach we have introduced by referring back to the requirements (boldface phrases) of \Cref{sec:purpose-limitation} and remarking several (additional) desirable properties, as well as limitations and opportunities for future work.

\subsection{Alternative Interpretations and Qualifications}
In \Cref{sec:gdpr} we have given an interpretation of the most relevant provisions of the GDPR that relate directly or indirectly to processing actions performed on personal data.
From this interpretation we have devised an ontology and an associated semantics that can be operationalised by a procedure that, effectively, computes an argument (if any) for the \textbf{lawfulness} of a particular processing action.
The (legal) \textbf{soundness} of this procedure is derived from the soundness of our interpretation of the GDPR.
The interpretation we have provided may be considered incorrect or \textbf{incomplete} as not all articles and recitals have been explicitly considered.
In the next paragraph on maintainability we will discuss the practical consequences when amendments are needed.
The interpretation is high-level and generic in the sense that it does not pertain to the details of any specific case, but can be re-used across the analysis of individual cases. 
The analysis of an individual case is completed by supplementing the reusable interpretation with qualifications recorded in the purpose graph. 
The details of the case need to be given on a case-by-case basis and amount to making qualifications about the applied legal basis and relevant purposes, including their specificity and compatibility. 
This way, \textbf{open-texture terms} are closed relative to the context of the case.

The details of individual cases matter as, for example, the GDPR describes \textbf{special categories of personal data} in \art{9}, prohibiting the processing of certain personal attributes, unless for specific purposes in specific situations.
For example, exceptions apply to specific purposes related to medicine, public health, and science.
A detailed analysis of purpose definition in the context of scientific research is provided by~\cite{10.1093/jlb/lsae001}.
The categories of special data, the prohibition of \artm{9}{1}, and its exceptions in \artmm{9}{2}{a-j} have not yet been introduced into our generic procedure.

\textbf{Other laws and regulations}, as well as contracts and organisational policies, may also apply to a case.
These additional sources of norms can be formalised independently and can be used in conjunction with our procedure for determining lawfulness, as is also briefly discussed in the next subsection.
Whilst we have focused on the GDPR in this paper, the purpose-concept is more widely applicable.
For example, purpose is also a core concept of the AI Act and of other national or international regulations.
The extent to which the concept of purpose and purpose graph, as defined in this paper, can be reused in formal interpretations of such regulations is left as future work.

In general, \textbf{different interpretations} of norms captured by natural language sources (regulation, contracts, etc.) may co-exist.
The policy administration point of \Cref{sec:integration} may be able to host multiple (versions of) interpretations of a legal text, deciding which is interpretation is to be applied on a per-request basis (e.g., depending on which controller applies to a request).

Interpretation of sources of norms is a complex task, even for experts.
Court cases and doctrine show that interpretation and qualification may be disputed ex-post and that disputes may in the end be decided by power arguments (e.g. a decision by the highest court).
Although we rely on an ex-ante formal interpretation and case-specific qualification, their transparency and explication enables the necessary scrutiny.

\subsection{Adaptability, Maintainability and Transparency}
By explicating both interpretations and qualifications in a specification separate from the (source text of the) implementation of a data processing system, we increase the \textbf{adaptability} and \textbf{maintainability} of the system.
The authorisations the system can generate are adjusted by modifying the policies maintained by a policy administration point, as described in \Cref{sec:technical-architecture}.
The applicable case-generic specification of the ontology and its semantics, the case-specific purpose graph, and any additional policies and policy information are determined on a per-request basis. 
As a result, either of these can be updated dynamically.

The case-generic specification may require modification to support alternative interpretations.
Although the formal interpretation is also administered in a policy administration point, the adaptation of the interpretation has some issues.
Firstly, our paper does not explicitly answer who is responsible for administering the interpretation.
Technically it is considered to be part of the system implementation and, as formulated here, the system does not support dynamic modifications to the interpretation applied (as it does for qualifications). 
Such dynamic updates are conceivable, and conceptually less arduous, because of the realised separation of concerns, but the details are left for future work.
Secondly, the nature of the adaptation may invalidate existing purpose graphs (e.g., if the \lstinline+legal-basis+ relation is modified).
And to ensure accountability, both the old and new versions of the interpretation and the purpose graphs need to be maintained persistently.

On the other hand, the formalisation of the interpretation is orthogonal to \textbf{additional policies} such as the local organisational policies of the controller that may be encoded in attribute-based or role-based access control policies.
Such policies can also be formalised in eFLINT (see~\cite{VANBINSBERGEN2022140}) and can be applied in conjunction with purpose graphs.
The eFLINT language has been designed with the required level of modularity of specifications in mind.
The purpose-based access control model of~\cite{byun2008purpose} describes roles and hierarchical data. 
Although these concepts can be defined and reasoned with in eFLINT, it remains to be seen to which extent the integration of hierarchical data, in particular, requires modifications to the code presented in \Cref{sec:implementation-semantics} or whether these concepts can be introduced separately.

The \textbf{transparency} of a system can be increased by communicating the rules applied to the relevant stakeholders.
The exact rules used to determine the lawfulness of individual processing actions can be persistently recorded, providing information as to why the processing was considered lawful \textit{at that time}.
This information can be communicated to relevant stakeholders such as privacy authorities and data subjects.
This way, as argued in \Cref{sec:integration,sec:ex-post-scenarios}, stakeholders are empowered to pinpoint concrete differences in interpretation or qualification, resulting in \textbf{explicit disagreements} that may be easier to resolve.
However, information is not equal to knowledge or understanding.
The technical information about the reasoning process that the system is capable of providing is likely to be too technical for legal experts in practical cases.
The ways in which this information can be used for providing explanations to different audiences (\textbf{explainability}) is left as future work.

\subsection{Applicability, User Experience and Automation}
\Cref{sec:integration} discusses how our approach can be integrated in novel or existing distributed (data exchange) systems that employ a form of access or usage control.
More research is required, however, to evaluate the \textbf{applicability} and \textbf{practicality} of the approach.
Towards this purpose we intend to apply our method to several real case scenarios by performing a legal analysis and technical assessment for each.
In particular, we intend to develop a prototype of a governance intermediary, following the design presented in \Cref{fig:intermediary-governed}, and experiment with the \textbf{user experience} of policy administration.
These experiments should reveal the preferred characteristics of the (programming) environment in which claims and qualifications about legal bases and purposes are made.
That is, which abstractions should be presented to the users and in which way? 
A possible answer is `User Story', as has been investigated by~\cite{10.1007/978-3-030-29238-6_1}.
And is policy administration a realistic ask from privacy or legal experts?
How does a user best document the decisions they have made?
The task of policy administration can potentially be simplified by integrating existing, standardised ontologies providing purpose definitions.  

In our approach, the actor making a request needs to state the purpose for performing a processing action called the \textbf{processing purpose}.
A technical question is how easy this requirement can be realised in existing implementations of access or usage control.
A possible solution, in which the processing purpose is determined as part of policy administration, has been discussed at the end of \Cref{sec:semantics}.
An interesting follow-up question is whether the processing purpose can be \textbf{determined} \textbf{automatically}.
In usage control, the processing activity itself can be monitored, potentially revealing information about the purpose of the processing or whether processing is being done for an additional, malign purpose.
Similarly, it may be possible to determine purpose by analysing sub-tasks of a processing activity, e.g., the tasks of a data analysis workflow.
In addition, by adopting a more structured representation of purposes and processing actions, it may be possible to (partially) automatically determine the `prerequisite of', `specific of' and `compatible with' relations. 
For example, regarding `compatible with', an approach could be to formalise purposes in terms of Boolean pre-conditions and post-conditions.
It is then possible to determine incompatibility between purposes by the presence of conflicts between the respective post-conditions of two purposes.
Similarly, by adopting fine-grained notions of data categories, purposes and processing actions, one might be able to assess the `minimality' of the processed data in comparison with the purpose for which a legal basis has been established. 
The user can be helped by the system generating edges in the purpose graph automatically whilst still leaving the final determination to the user.

Determining the purpose of a piece of software is difficult.
As software artefacts themselves do not possess intentions, nor purposes they seek to fulfill, we may only have access to what the software could \textit{possibly} do with the data or what people state about the software.
The purposes of the designer or user of software cannot be accessed directly and people may have reasons to pretend.
Even when the designer or user is clear and truthful about their purposes, the software may contain bugs they are not aware of. 
In the case the source code of the software is available, we may be able to determine what the code could possibly do with any personal data through code inspection. 
This approach involves a worst-case approximation, as any malign code may never be reached in the specific context in which the software is applied. 
If the source code is not available, we may have to resort to black-box testing.
This approach runs the risk of being incomplete, as it may not be possible to anticipate all possible runs of the software a priori.

\subsection{Norm Representation and Operationalisation}
To represent knowledge and norms we have used eFLINT, a domain-specific \textbf{norm specification language} designed for specifying and reasoning with norms as found in laws, regulations, and contracts.
Other software languages could have been used instead; our approach does not make strong assumptions that may impact the ease of integration or the \textbf{operationalisation} of the semantics.
However, we argue for the use of logical programming languages, and declarative rule-based languages more generally, as the application of rules and logical inference has the potential to increase transparency and explainability. 
The provided semantics is in the form of first-order logic and does not require the use of negation.
In fact, any language adhering to description logic with a classifier could have been used.
As such, Prolog, Datalog and Clingo are all viable alternatives, as well as general-purpose and database programming languages.

Such implementations would (natively) lack the possibility to reason about the normative positions of actors (such as permissions, prohibitions, obligations, powers, and duties).
%
%involved that by performing certain institutional acts change those normative positions (both for themselves as for other agents). An in depth discussion of the differences between eFlint and potential alternative knowledge representation and reasoning alternatives will be addressed in a separate paper.
%
Our choice for eFLINT is further motivated by the desire to combine lawfulness according to the GDPR with additional normative specifications, e.g., of data processing agreements or data sharing conditions, that may assign normative positions to actors. 
Other norm specifications languages based on rules or logic programming have been proposed, such as InstAL~(\cite{padget2016instal}), Symboleo~(\cite{symboleo2020}) and various logics (such as ~\cite{herrestad1991,nute2003,governatori2004,jones_sergot_1996}).
The mentioned languages display subtle or more fundamental differences in the chosen norm representation and scope of application.
For example, eFLINT and Symboleo are both based on Hohfeldian principles (see~\cite{hohfeldFundamentalLegalConceptions1913,hohfeld1917fundamental}), but Symboleo is designed specifically for contracts, whereas eFLINT is designed for norms more generally. 
Norm representation and normative reasoning have been studied for some decades and in early work norm representation in first order logic with an implementation in Prolog were quite common (see, for example, \cite{10.1145/5689.5920}).
For a more detailed comparison between eFLINT and normative alternatives, the reader is referred to~\cite{eflint}.

An implementation of our approach in a language supporting a form of search, such as Prolog or Clingo, can potentially be used to search for (completions of) lawfulness arguments.
That is, the system can help finding a legal basis given a complete purpose specification or can help identifying where a compatibility assessment between purposes needs to be made for a particular legal basis, increasing usability.
%

% In this paper we present an approach to formalize purposes in a fine-grained manner.
% %
% This provides data controllers with a system that enables them to fulfill both a legal requirement as well as that it allows them to overcome the limitations of systems merely focused on access control, as the GDPR limits data processing including access to data, hence usage control.
% %
% Being able to control data usage in not only essential for ensuring compliance with the purpose limitation principle in the GDPR. Purpose is also a core concept in the AI Act and other (national) regulations and policies.
% %

%
% Our approach offers valuable benefits to policy drafters to design GDPR-compliant purpose-based policies.
% %
% After a concise summary of the purpose limitation principle and related work in purpose-based usage control we have demonstrated our approach using a fictional case.
% %
% We introduced an action matching algorithm leveraging the eFLINT reasoner to align data usage query actions with permissible ones.%
% While our approach is a promising initial step toward GDPR-compliant purpose-based usage control, challenges remain.
% %
% Future research directions include automating compatible purpose validation, utilizing legal ontologies for legitimacy checks, and involving legal experts to  identify additional relationships between purposes.
% %
% Furthermore, adding other attributes such as roles, relationships between parties involved could result in complex policies that satisfy different domain-specific needs. 

\section{Conclusion}
\label{sec:conclusion}
We have presented an ontology and semantics for automated normative reasoning about the lawfulness of data processing according to the GDPR.
The ontology and semantics are case-generic, and can be used across use cases and within multiple types of data processing systems.
By separating the rules formalising the norms from the implementation of a system, the transparency, adaptability and maintainability of the system are increased.
General guidelines to integrating the proposed normative reasoning within (existing) distributed data processing systems have been described.

To make decisions on individual cases, case-specific qualifications are made by specific users in the system (e.g. controller's legal basis and subject's consent).
The system ensures processing actions are only authorised when a valid argument for the lawfulness of the processing action can be constructed from a given set of qualifications.
By persistently recording the qualifications, decisions made within the system are made accountable, and can be used to offer explanations to various stakeholders involved.
GDPR compliance can thus be enforced both through ex-ante procedures in access or usage control systems as well as in ex-post auditing scenarios. 

The approach proposed in this paper decreases the risk of GDPR non-compliance and reduces the effort of demonstrating compliance and satisfying the accountability requirements of the GDPR.
Moreover, policy makers or legal/privacy experts can directly influence system behaviour through case-specific qualifications and no longer have to rely on a system administrator or software expert to configure or program their system correctly.
The presented approach reduces effort by providing a case-generic interpretation of the GDPR and automating reasoning whilst retaining legal soundness by leaving crucial qualifications (e.g., purpose-determination and closing open-texture terms) to expert users.

This paper focusses on the purpose-limitation principle of the GDPR and demonstrates the core concepts of our approach to GDPR-based normative reasoning in distributed data processing systems.
A more complete treatment of the GDPR, e.g., also considering `joint controllership' and special categories of personal data, are left as future work.
The potential impact of the suggested approach regarding user experience and efficiency gains remain to be evaluated in real use cases and practical systems.

% This paper presents an approach to formalize purposes in a fine-grained manner within usage control systems, ensuring compliance with the purpose limitation principle. With our approach we hope to offer valuable benefits to policy authors to design GDPR-compliant purpose-based policies. We demonstrated our approach using a store use case, for which we introduced an action matching algorithm leveraging the eFLINT reasoner to align usage query actions with permissible ones. Our approach is a promising initial step toward GDPR-compliant purpose-based usage control. Future research directions include automating compatible purpose validation, utilizing legal ontologies for legitimacy checks, and involving legal experts to identify more sophisticated  relationships between purposes.

\section*{Acknowledgements}
The authors would like to thank Jeroen Jongenelen for the description of the KPN wiretapping case.
This research is partially funded as part of the ``Data sharing through a data intermediary'' project within the Responsible Digital Transformations programme of the University of Amsterdam and the AMdEX-DMI project supported by the Dutch Metropolitan Innovations ecosystem for smart and sustainable cities, made possible by the Nationaal Groeifonds.


\bibliography{sn-bibliography}% common bib file

\begin{thebibliography}{}
\providecommand{\doi}[1]{\url{https://doi.org/#1}}
\bibcommenthead

\bibitem[\protect\citeauthoryear{Ashley, Hada, Karjoth, Powers, and Schunter}{Ashley et~al.}{2003}]{ashley2003enterprise}
Ashley, P., S.~Hada, G.~Karjoth, C.~Powers, and M.~Schunter. 2003.
\newblock Enterprise privacy authorization language (epal).
\newblock {\em IBM Research\/}~30: 31 .

\bibitem[\protect\citeauthoryear{Baramashetru, Tapia~Tarifa, and Owe}{Baramashetru et~al.}{2024}]{10.1007/978-3-031-57978-3_4}
Baramashetru, C.P., S.L. Tapia~Tarifa, and O.~Owe 2024.
\newblock Assuring gdpr conformance through language-based compliance.
\newblock In F.~Bieker, S.~de~Conca, N.~Gruschka, M.~Jensen, and I.~Schiering (Eds.), {\em Privacy and Identity Management. Sharing in a Digital World}, Cham, pp.\  46--63. Springer Nature Switzerland.

\bibitem[\protect\citeauthoryear{Bartolini, Daoudagh, Lenzini, and Marchetti}{Bartolini et~al.}{2019}]{10.1007/978-3-030-29238-6_1}
Bartolini, C., S.~Daoudagh, G.~Lenzini, and E.~Marchetti 2019.
\newblock Gdpr-based user stories in the access control perspective.
\newblock In M.~Piattini, P.~Rupino~da Cunha, I.~Garc{\'i}a Rodr{\'i}guez~de Guzm{\'a}n, and R.~P{\'e}rez-Castillo (Eds.), {\em Quality of Information and Communications Technology}, Cham, pp.\  3--17. Springer International Publishing.

\bibitem[\protect\citeauthoryear{Basin, Debois, and Hildebrandt}{Basin et~al.}{2018}]{basin2018purpose}
Basin, D., S.~Debois, and T.~Hildebrandt 2018.
\newblock {On purpose and by necessity: compliance under the GDPR}.
\newblock In {\em Financial Cryptography and Data Security: 22nd International Conference, FC 2018, Nieuwpoort, Cura{\c{c}}ao, February 26--March 2, 2018, Revised Selected Papers 22}, pp.\  20--37. Springer.

\bibitem[\protect\citeauthoryear{Becker, Chokoshvili, Thorogood, Dove, Molnár-Gábor, Ziaka, Tzortzatou-Nanopoulou, and Comandè}{Becker et~al.}{2024}]{10.1093/jlb/lsae001}
Becker, R., D.~Chokoshvili, A.~Thorogood, E.S. Dove, F.~Molnár-Gábor, A.~Ziaka, O.~Tzortzatou-Nanopoulou, and G.~Comandè. 2024, 02.
\newblock {Purpose definition as a crucial step for determining the legal basis under the GDPR: implications for scientific research}.
\newblock {\em Journal of Law and the Biosciences\/}~{\em 11\/}(1): lsae001.
\newblock \doi{10.1093/jlb/lsae001} .

\bibitem[\protect\citeauthoryear{Byun, Bertino, and Li}{Byun et~al.}{2005}]{byun2005purpose}
Byun, J.W., E.~Bertino, and N.~Li 2005.
\newblock Purpose based access control of complex data for privacy protection.
\newblock In {\em Proceedings of the tenth ACM symposium on Access control models and technologies}, pp.\  102--110.

\bibitem[\protect\citeauthoryear{Byun and Li}{Byun and Li}{2008}]{byun2008purpose}
Byun, J.W. and N.~Li. 2008.
\newblock Purpose based access control for privacy protection in relational database systems.
\newblock {\em The VLDB Journal\/}~17: 603--619 .

\bibitem[\protect\citeauthoryear{Ceri, Gottlob, and Tanca}{Ceri et~al.}{1989}]{datalog}
Ceri, S., G.~Gottlob, and L.~Tanca. 1989.
\newblock What you always wanted to know about datalog (and never dared to ask).
\newblock {\em IEEE Transactions on Knowledge and Data Engineering\/}~{\em 1\/}(1): 146--166.
\newblock \doi{10.1109/69.43410} .

\bibitem[\protect\citeauthoryear{Chen, Han, Zhang, You, and Zheng}{Chen et~al.}{2023}]{CHEN2023102938}
Chen, J., P.~Han, Y.~Zhang, T.~You, and P.~Zheng. 2023.
\newblock Scheduling energy consumption-constrained workflows in heterogeneous multi-processor embedded systems.
\newblock {\em Journal of Systems Architecture\/}~142: 102938.
\newblock \doi{10.1016/j.sysarc.2023.102938} .

\bibitem[\protect\citeauthoryear{Chowdhury, Chen, Niu, Li, and Bertino}{Chowdhury et~al.}{2012}]{chowdhury2012}
Chowdhury, O., H.~Chen, J.~Niu, N.~Li, and E.~Bertino 2012.
\newblock {On XACML's adequacy to specify and to enforce HIPAA}.
\newblock In {\em Proceedings of the 3rd USENIX Conference on Health Security and Privacy}, HealthSec'12, USA, pp.\ ~11. USENIX Association.

\bibitem[\protect\citeauthoryear{Cohen, Crampton, Gagarin, Gutin, and Jones}{Cohen et~al.}{2014}]{Cohen2014}
Cohen, D., J.~Crampton, A.~Gagarin, G.~Gutin, and M.~Jones. 2014, sep.
\newblock Iterative plan construction for the workflow satisfiability problem.
\newblock {\em J. Artif. Int. Res.\/}~{\em 51\/}(1): 555–577 .

\bibitem[\protect\citeauthoryear{{Council of the EU}}{{Council of the EU}}{2016}]{gdpr}
{Council of the EU}. 2016.
\newblock {General Data Protection Regulation}.
\newblock {\em Official Journal of the European Union\/}~59 .

\bibitem[\protect\citeauthoryear{{Council of the EU}}{{Council of the EU}}{2020}]{eu-data-strategy}
{Council of the EU}. 2020.
\newblock {A European strategy for data}.
\newblock {\em Official Journal of the European Union\/} .

\bibitem[\protect\citeauthoryear{De~Masellis, Ghidini, and Ranise}{De~Masellis et~al.}{2015}]{de2015declarative}
De~Masellis, R., C.~Ghidini, and S.~Ranise 2015.
\newblock A declarative framework for specifying and enforcing purpose-aware policies.
\newblock In {\em Security and Trust Management: 11th International Workshop, STM 2015, Vienna, Austria, September 21-22, 2015, Proceedings 11}, pp.\  55--71. Springer.

\bibitem[\protect\citeauthoryear{Degeler and Lazovik}{Degeler and Lazovik}{2014}]{Degeler2014}
Degeler, V. and A.~Lazovik. 2014.
\newblock Dynamic constraint satisfaction with space reduction in smart environments.
\newblock {\em International Journal on Artificial Intelligence Tools\/}~{\em 23\/}(06): 1460027.
\newblock \doi{10.1142/S0218213014600276} .

\bibitem[\protect\citeauthoryear{Gabbay, Hogger, Robinson, and Siekmann}{Gabbay et~al.}{1994}]{10.1093/oso/9780198537465.001.0001}
Gabbay, D.M., C.J. Hogger, J.A. Robinson, and J.~Siekmann. 1994, 03.
\newblock {\em {Handbook of Logic in Artificial Intelligence and Logic Programming}}.
\newblock Oxford University Press.

\bibitem[\protect\citeauthoryear{Governatori, Maher, Antoniou, and Billington}{Governatori et~al.}{2004}]{governatori2004}
Governatori, G., M.~Maher, G.~Antoniou, and D.~Billington. 2004, 10.
\newblock {Argumentation Semantics for Defeasible Logic}.
\newblock {\em Journal of Logic and Computation\/}~{\em 14\/}(5): 675--702.
\newblock \doi{10.1093/logcom/14.5.675} .

\bibitem[\protect\citeauthoryear{Hariri, Ibrahim, Alangot, Bandopadhyay, La~Marra, Rosetti, Joumaa, and Dimitrakos}{Hariri et~al.}{2023}]{Hariri2023}
Hariri, A., A.~Ibrahim, B.~Alangot, S.~Bandopadhyay, A.~La~Marra, A.~Rosetti, H.~Joumaa, and T.~Dimitrakos 2023.
\newblock {\em UCON+: Comprehensive Model, Architecture and Implementation for Usage Control and Continuous Authorization}, pp.\  209--226.
\newblock Cham: Springer International Publishing.

\bibitem[\protect\citeauthoryear{Herrestad}{Herrestad}{1993}]{herrestad1991}
Herrestad, H. 1993.
\newblock Norms and formalization.
\newblock In {\em Proceedings of the 3th International Conference on Artificial Intelligence and Law}, ICAIL 1993, pp.\  175--184. ACM.

\bibitem[\protect\citeauthoryear{Hohfeld}{Hohfeld}{1913}]{hohfeldFundamentalLegalConceptions1913}
Hohfeld, W.N. 1913, 11.
\newblock Some {{Fundamental Legal Conceptions}} as {{Applied}} in {{Judicial Reasoning}}.
\newblock ~{\em 23\/}(1): 16--16.
\newblock \doi{10.2307/785533} .

\bibitem[\protect\citeauthoryear{Hohfeld}{Hohfeld}{1917}]{hohfeld1917fundamental}
Hohfeld, W.N. 1917.
\newblock Fundamental legal conceptions as applied in judicial reasoning.
\newblock {\em The Yale Law Journal\/}~{\em 26\/}(8): 710--770 .

\bibitem[\protect\citeauthoryear{Hu, Kuhn, Ferraiolo, and Voas}{Hu et~al.}{2015}]{abac}
Hu, V.C., D.R. Kuhn, D.F. Ferraiolo, and J.~Voas. 2015.
\newblock Attribute-based access control.
\newblock {\em Computer\/}~{\em 48\/}(2): 85--88.
\newblock \doi{10.1109/MC.2015.33} .

\bibitem[\protect\citeauthoryear{Iyilade and Vassileva}{Iyilade and Vassileva}{2014}]{iyilade2014p2u}
Iyilade, J. and J.~Vassileva 2014.
\newblock {P2U: a privacy policy specification language for secondary data sharing and usage}.
\newblock In {\em 2014 IEEE Security and Privacy Workshops}, pp.\  18--22. IEEE.

\bibitem[\protect\citeauthoryear{Jafari, Safavi-Naini, and Sheppard}{Jafari et~al.}{2009}]{jafari2009enforcing}
Jafari, M., R.~Safavi-Naini, and N.P. Sheppard 2009.
\newblock Enforcing purpose of use via workflows.
\newblock In {\em Proceedings of the 8th ACM Workshop on Privacy in the Electronic Society}, pp.\  113--116.

\bibitem[\protect\citeauthoryear{Jones and Sergot}{Jones and Sergot}{1996}]{jones_sergot_1996}
Jones, A. and M.~Sergot. 1996, 06.
\newblock {A Formal Characterisation of Institutionalised Power}.
\newblock {\em Logic Journal of the IGPL\/}~{\em 4\/}(3): 427--443.
\newblock \doi{10.1093/jigpal/4.3.427} .

\bibitem[\protect\citeauthoryear{Jung and D{\"o}rr}{Jung and D{\"o}rr}{2022}]{Jung2022}
Jung, C. and J.~D{\"o}rr 2022.
\newblock {\em Data Usage Control}, pp.\  129--146.
\newblock Cham: Springer International Publishing.

\bibitem[\protect\citeauthoryear{Kassem, Valkering, Belloum, and Grosso}{Kassem et~al.}{2021}]{Kassem2021}
Kassem, J.A., O.~Valkering, A.~Belloum, and P.~Grosso 2021.
\newblock Epi framework: Approach for traffic redirection through containerised network functions.
\newblock In {\em 2021 IEEE 17th International Conference on eScience (eScience)}, pp.\  80--89.

\bibitem[\protect\citeauthoryear{Kebede, Sileno, and Van~Engers}{Kebede et~al.}{2021}]{10.1007/978-3-030-89811-3_4}
Kebede, M.G., G.~Sileno, and T.~Van~Engers 2021.
\newblock {A Critical Reflection on ODRL}.
\newblock In V.~Rodr{\'i}guez-Doncel, M.~Palmirani, M.~Araszkiewicz, P.~Casanovas, U.~Pagallo, and G.~Sartor (Eds.), {\em AI Approaches to the Complexity of Legal Systems XI-XII}, Cham, pp.\  48--61. Springer International Publishing.

\bibitem[\protect\citeauthoryear{Kowalski, D{\'a}vila, Sartor, and Calejo}{Kowalski et~al.}{2023}]{Kowalski2023}
Kowalski, R., J.~D{\'a}vila, G.~Sartor, and M.~Calejo 2023.
\newblock {\em Logical English for Law and Education}, pp.\  287--299.
\newblock Cham: Springer Nature Switzerland.

\bibitem[\protect\citeauthoryear{Maier, Tekle, Kifer, and Warren}{Maier et~al.}{2018}]{datalog-book}
Maier, D., K.T. Tekle, M.~Kifer, and D.S. Warren. 2018.
\newblock Datalog: concepts, history, and outlook, In {\em Declarative Logic Programming: Theory, Systems, and Applications},  eds. Kifer, M. and Y.A. Liu, Volume~20 of {\em {ACM} Books},  3--100. {ACM} / Morgan {\&} Claypool.
\newblock \doi{10.1145/3191315.3191317}.

\bibitem[\protect\citeauthoryear{Nute}{Nute}{2003}]{nute2003}
Nute, D. 2003.
\newblock Defeasible logic.
\newblock In O.~Bartenstein, U.~Geske, M.~Hannebauer, and O.~Yoshie (Eds.), {\em Web Knowledge Management and Decision Support}, Berlin, Heidelberg, pp.\  151--169. Springer Berlin Heidelberg.

\bibitem[\protect\citeauthoryear{Otto, Steinbuss, Teuscher, and Lohmann}{Otto et~al.}{2019}]{idsa-ram}
Otto, B., S.~Steinbuss, A.~Teuscher, and S.~Lohmann. 2019, 4.
\newblock {IDS Reference Architecture Model -- version 3.0}.

\bibitem[\protect\citeauthoryear{Padget, Elakehal, Li, and De~Vos}{Padget et~al.}{2016}]{padget2016instal}
Padget, J., E.E. Elakehal, T.~Li, and M.~De~Vos. 2016.
\newblock {InstAL}: An institutional action language, {\em Social coordination frameworks for social technical systems},  101--124. Springer.
\newblock \doi{10.1007/978-3-319-33570-4_6}.

\bibitem[\protect\citeauthoryear{Pallas, Ulbricht, Tai, Peikert, Reppenhagen, Wenzel, Wille, and Wolf}{Pallas et~al.}{2020}]{pallas2020towards}
Pallas, F., M.R. Ulbricht, S.~Tai, T.~Peikert, M.~Reppenhagen, D.~Wenzel, P.~Wille, and K.~Wolf 2020.
\newblock Towards application-layer purpose-based access control.
\newblock In {\em Proceedings of the 35th Annual ACM Symposium on Applied Computing}, pp.\  1288--1296.

\bibitem[\protect\citeauthoryear{Petkovi{\'c}, Prandi, and Zannone}{Petkovi{\'c} et~al.}{2011}]{petkovic2011purpose}
Petkovi{\'c}, M., D.~Prandi, and N.~Zannone 2011.
\newblock Purpose control: Did you process the data for the intended purpose?
\newblock In {\em Secure Data Management: 8th VLDB Workshop, SDM 2011, Seattle, WA, USA, September 2, 2011, Proceedings 8}, pp.\  145--168. Springer.

\bibitem[\protect\citeauthoryear{Pretschner, Hilty, and Basin}{Pretschner et~al.}{2006}]{Pretschner2006}
Pretschner, A., M.~Hilty, and D.~Basin. 2006, sep.
\newblock Distributed usage control.
\newblock {\em Commun. ACM\/}~{\em 49\/}(9): 39–44.
\newblock \doi{10.1145/1151030.1151053} .

\bibitem[\protect\citeauthoryear{Saewong and Rajkumar}{Saewong and Rajkumar}{1999}]{Saewong1999}
Saewong, S. and R.~Rajkumar 1999.
\newblock Cooperative scheduling of multiple resources.
\newblock In {\em Proceedings 20th IEEE Real-Time Systems Symposium (Cat. No.99CB37054)}, pp.\  90--101.

\bibitem[\protect\citeauthoryear{Sandhu, Coyne, Feinstein, and Youman}{Sandhu et~al.}{1996}]{rbac}
Sandhu, R., E.~Coyne, H.~Feinstein, and C.~Youman. 1996.
\newblock Role-based access control models.
\newblock {\em Computer\/}~{\em 29\/}(2): 38--47.
\newblock \doi{10.1109/2.485845} .

\bibitem[\protect\citeauthoryear{Sandhu and Park}{Sandhu and Park}{2003}]{10.1007/978-3-540-45215-7_2}
Sandhu, R. and J.~Park 2003.
\newblock Usage control: A vision for next generation access control.
\newblock In V.~Gorodetsky, L.~Popyack, and V.~Skormin (Eds.), {\em Computer Network Security}, Berlin, Heidelberg, pp.\  17--31. Springer Berlin Heidelberg.

\bibitem[\protect\citeauthoryear{Satoh}{Satoh}{2023}]{Satoh2023}
Satoh, K. 2023.
\newblock {\em PROLEG: Practical Legal Reasoning System}, pp.\  277--283.
\newblock Cham: Springer Nature Switzerland.

\bibitem[\protect\citeauthoryear{Sergot, Sadri, Kowalski, Kriwaczek, Hammond, and Cory}{Sergot et~al.}{1986}]{10.1145/5689.5920}
Sergot, M.J., F.~Sadri, R.A. Kowalski, F.~Kriwaczek, P.~Hammond, and H.T. Cory. 1986, May.
\newblock The british nationality act as a logic program.
\newblock {\em Commun. ACM\/}~{\em 29\/}(5): 370–386.
\newblock \doi{10.1145/5689.5920} .

\bibitem[\protect\citeauthoryear{Shakeri, Veen, and Grosso}{Shakeri et~al.}{2020}]{shakeri2020}
Shakeri, S., L.~Veen, and P.~Grosso 2020.
\newblock Evaluation of container overlays for secure data sharing.
\newblock In {\em 2020 IEEE 45th LCN Symposium on Emerging Topics in Networking (LCN Symposium)}, pp.\  99--108.

\bibitem[\protect\citeauthoryear{Sharifi, Parvizimosaed, Amyot, Logrippo, and Mylopoulos}{Sharifi et~al.}{2020}]{symboleo2020}
Sharifi, S., A.~Parvizimosaed, D.~Amyot, L.~Logrippo, and J.~Mylopoulos 2020.
\newblock Symboleo: Towards a {S}pecification {L}anguage for {L}egal {C}ontracts.
\newblock In T.~D. Breaux, A.~Zisman, S.~Fricker, and M.~Glinz (Eds.), {\em 28th {IEEE} International Requirements Engineering Conference, {RE} 2020, August 31 - September 4, 2020}, pp.\  364--369. {IEEE}.

\bibitem[\protect\citeauthoryear{Spiekermann}{Spiekermann}{2019}]{Spiekermann2019}
Spiekermann, M. 2019, 7.
\newblock {Data Marketplaces: Trends and Monetisation of Data Goods}.
\newblock {\em Intereconomics\/}.
\newblock \doi{10.1007/s10272-019-0826-z} .

\bibitem[\protect\citeauthoryear{Tschantz, Datta, and Wing}{Tschantz et~al.}{2012}]{tschantz2012formalizing}
Tschantz, M.C., A.~Datta, and J.M. Wing 2012.
\newblock Formalizing and enforcing purpose restrictions in privacy policies.
\newblock In {\em 2012 IEEE Symposium on Security and Privacy}, pp.\  176--190. IEEE.

\bibitem[\protect\citeauthoryear{van Binsbergen, Kebede, Baugh, van Engers, and van Vuurden}{van Binsbergen et~al.}{2022}]{VANBINSBERGEN2022140}
van Binsbergen, L.T., M.G. Kebede, J.~Baugh, T.~van Engers, and D.G. van Vuurden. 2022.
\newblock Dynamic generation of access control policies from social policies.
\newblock {\em Procedia Computer Science\/}~198: 140--147.
\newblock \doi{10.1016/j.procs.2021.12.221} .

\bibitem[\protect\citeauthoryear{van Binsbergen, Liu, van Doesburg, and van Engers}{van Binsbergen et~al.}{2020}]{eflint}
van Binsbergen, L.T., L.~Liu, R.~van Doesburg, and T.~van Engers 2020.
\newblock {eFLINT}: A {D}omain-{S}pecific {L}anguage for {E}xecutable {N}orm {S}pecifications.
\newblock In {\em Proceedings of the 19th ACM SIGPLAN International Conference on Generative Programming: Concepts and Experiences}, GPCE 2020, pp.\  124--–136. ACM.

\bibitem[\protect\citeauthoryear{van Binsbergen, Oost-Rosengren, Schreijer, Dijkstra, and van Dijk}{van Binsbergen et~al.}{2024}]{amdex-ra}
van Binsbergen, L.T., M.~Oost-Rosengren, H.~Schreijer, F.~Dijkstra, and T.~van Dijk. 2024, 2.
\newblock {\em AMdEX Reference Architecture -- version 1.0.0}.

\bibitem[\protect\citeauthoryear{Veen, Shakeri, and Grosso}{Veen et~al.}{2022}]{mahiru}
Veen, L.E., S.~Shakeri, and P.~Grosso. 2022.
\newblock Mahiru: a federated, policy-driven data processing and exchange system.
\newblock {\em CoRR\/}~abs/2210.17155.
\newblock \doi{10.48550/ARXIV.2210.17155}.
\newblock {\href{https://arxiv.org/abs/2210.17155}{{2210.17155}}} .

\bibitem[\protect\citeauthoryear{Warren}{Warren}{2023}]{Warren2023}
Warren, D.S. 2023.
\newblock {\em Introduction to Prolog}, pp.\  3--19.
\newblock Cham: Springer Nature Switzerland.

\bibitem[\protect\citeauthoryear{Zhang, Fan, Zhou, and Zhou}{Zhang et~al.}{2019}]{zhang2019purpose}
Zhang, F., X.~Fan, W.~Zhou, and P.~Zhou. 2019.
\newblock Purpose-based access policy on provenance and data algebra.
\newblock {\em arXiv preprint arXiv:1912.00445\/} .

\bibitem[\protect\citeauthoryear{Zhang, Cushing, Gommans, De~Laat, and Grosso}{Zhang et~al.}{2019}]{zhang2019}
Zhang, L., R.~Cushing, L.~Gommans, C.~De~Laat, and P.~Grosso. 2019.
\newblock Modeling of collaboration archetypes in digital market places.
\newblock {\em IEEE Access\/}~7: 102689--102700.
\newblock \doi{10.1109/ACCESS.2019.2931762} .

\end{thebibliography}

\begin{appendices}

%\section{Section title of first appendix}\label{secA1}




\end{appendices}

\end{document}